%% file: Thesis.tex
\documentclass[a4paper,12pt]{scrbook}

\usepackage[english]{babel}

\usepackage{amsmath}
\usepackage{amssymb}
\usepackage{amsthm}
\usepackage{graphicx}
\usepackage{subfig}
\usepackage{tabularx}
\usepackage{textcomp}
\usepackage{url}
\usepackage[numbers]{natbib}
\usepackage{rotating}
\usepackage{tikz}
\usetikzlibrary{plotmarks}
\usepackage{hyperref}
\hypersetup{colorlinks=false}

\newcommand{\autor}{Philipp Ewerling}
\newcommand{\titel}{A novel processing pipeline for optical multi-touch surfaces}
\newcommand{\typ}{MSc Thesis}
\newcommand{\betreuer}{Prof. Dr. Bernd Fr\"ohlich}
\newcommand{\datum}{29. Februar 2012}
\newcommand{\datumenglisch}{29 February 2012}

\begin{document}
\frontmatter
\begin{titlepage}
\begin{flushleft}
\large
Bauhaus-Universit\"at Weimar\\
Fakult\"at Medien\\
Studiengang Mediensysteme\\
\end{flushleft}

\vspace{3cm}

\begin{center}
\textsf{ 
\huge\textbf{\titel}\\
\vspace{2cm}
\huge\textbf{\typ}\\
\vspace{4cm}
}
\end{center}
\begin{center}
\Large
\textbf{\autor}\\
\large
\vspace*{1cm}
1. Gutachter: \betreuer\\

\vspace*{1cm}
\large
Datum der Abgabe: \datum\\
\end{center}
\end{titlepage}

\cleardoublepage

\chapter*{Author's Declaration}

Unless otherwise indicated in the text or references, this thesis is entirely the product of my own scholarly work. Any inaccuracies of fact or faults in reasoning are my own and accordingly I take full responsibility. This thesis has not been submitted either in whole or part, for a degree at this or any other university or institution. This is to certify that the printed version is equivalent to the submitted electronic one.\\
\\
\\
\\
\\
Weimar, \datumenglisch
\\

\cleardoublepage

\tableofcontents

\mainmatter

\input{./Content/Introduction}
\input{./Content/Chapter_1}

\input{./Content/Chapter_2}

\input{./Content/Chapter_3}
\input{./Content/Chapter_4}

\appendix
\cleardoublepage
\phantomsection
\addcontentsline{toc}{chapter}{List of Figures}

\listoffigures

\cleardoublepage
\phantomsection
\addcontentsline{toc}{chapter}{Bibliography}


\end{document}

%% file: Content/Introduction.tex
\chapter{Introduction}
Multi-touch technology has recently emerged as one of the most sought after input technologies. The presentation of the iPhone and the Microsoft Surface, both in 2007, have received considerable attention and touch-sensing technology has since then become a de facto standard in consumer electronics. Moreover the presentation of a low-cost multi-touch prototype by Han in 2005 based on the principle of \textit{frustrated total internal reflection} has made the required sensing technology available to a large community that is continuously driving the development of new multi-touch applications.

In the course of time multi-touch researchers and developers have come up with an ever growing number of gestures used in multi-touch devices, however they generally have a common limitation. Most devices assume that all simultaneous touches belong to the same gesture, hence support only a single user. While this is without question true for small form factor devices like smartphones, things aren't as straightforward for large touch screens that can accommodate more than one user. One approach, as used in the Microsoft Surface, is to separate multiple simultaneous touches based on the touched object and to evaluate gestures for each object individually. Another approach is to define public and private areas on the touch-sensitive surface granting each user a private workspace. Albeit the aptitude of these approaches they are constraint by the data provided from the sensing technology.

Technologies based on resistive and acoustic approaches are by design limited to sensing direct surface contact only, while capacitive technology allows to a certain extend also the sensing of objects in proximity. Apart from the MERL Diamond Touch\footnote{see page \pageref{section:tech_diamondtouch}} which can uniquely identify the interacting user for each touch point, these technologies reveal solely the position of a surface contact and possibly its size and shape. In contrast technologies based on optical sensing are able to provide a much richer representation of what happens on and beyond the touch-sensitive surface. Although this entails a significantly higher processing performance due to the increased complexity, it opens the possibility of providing contextual information for each surface contact. Hence one might be able to define whether simultaneous touch points originate from the same or different hands. Further information might include whether a finger touch stems from a left or right hand and might even help to identify users given the orientation of the hand. 

Yet surprisingly little research exists on how to extract these properties from the camera image. On the one hand processing in most prototypes is limited to a rather simple surface contact detection algorithm ignoring all these aforementioned factors. On the other hand the existing approaches all try to establish spatial relationships solely from the previously revealed contact position and shape. However no integrated processing pipeline exists that combines those features and reveals surface touches together with their spatial relationships.

This thesis proposes a novel processing pipeline for optical multi-touch surfaces that fully incorporates these features. Based on the \textit{Maximally Stable Extremal Regions} algorithm that has previously been mainly used in stereo image matching and visual tracking, processing steps will be described that by design reveal surface contacts as well as their spatial relationships. Taking advantage of this structure finger touches can not only be attributed to different hands but in presence of all five fingers the algorithm even performs a hand and finger registration.

However before further elaborating the processing pipeline, the next chapter will first present a survey of the different sensing technologies and detail the most common processing steps for prototypes based on optical sensing. Afterwards the multi-touch prototype and the novel processing pipeline will be thoroughly described in chapter \ref{chapter:pipeline}. Chapter \ref{chapter:evaluation} follows with an evaluation of both performance and accuracy of the proposed algorithms, while in chapter \ref{chapter:conclusion} a conclusion and an outlook on potential future work will be given.

%% file: Content/Chapter_1.tex
\chapter{Existing Multi-Touch Technologies}\label{chapter:related_work}
Although multi-touch devices only recently caught the attention of a wider public with the proliferation of touch sensing technologies in smartphones and tablet computers, it actually has a very long history ranging back until the early 1980s. Bill Buxton, himself an important contributor to the multi-touch research, names the \textit{Flexible Machine Interface} \cite{metha1982flexible} from 1982 as the first multi-touch system in his much-cited review on touch-sensitive devices \cite{buxton2011multitouch}. The prototype consisted of a panel of frosted glass on which the finger's shadows appeared as black spots with their size increasing the more the user touched the panel. It is the earliest prototype where a camera mounted on the underside of the tabletop was used to capture the surface. Further examples of multi-touch devices from the early stages, this time based on capacitive sensing, have been presented by Bob Boie in 1984 \cite{boie1984capacitive} and one year later by \citeauthor{lee1985multi} \cite{lee1985multi}. 

Since these early prototypes many other devices have been developed that rely on a wide range of different sensing technologies. Hence, this chapter will focus in a first section on that variety of sensing technologies, including widely known techniques based on capacitive and optical sensing as well as lesser known ones based on acoustic and thermal sensing. In a second section the commonly used processing steps required to reveal finger touches in a diffuse back-illumination setup, the same as the one used in the prototype presented in the next chapter, will be described. This pipeline will be considered the de facto standard when introducing the novel processing pipeline.

\section{Sensing Technologies}
All existing sensing technologies have a number of advantages and disadvantages when it comes to sensing finger touch. The following classification of technologies loosely follows the one proposed in \cite{schöning2008multi}, however adds missing technologies and examples wherever required. This section will outline the functioning of resistive, capacitive, acoustic and optical sensing and present relevant prototypes.

\subsection{Resistive Sensing}
\begin{figure}[t]
	\centering
    \includegraphics[width=0.9\textwidth]{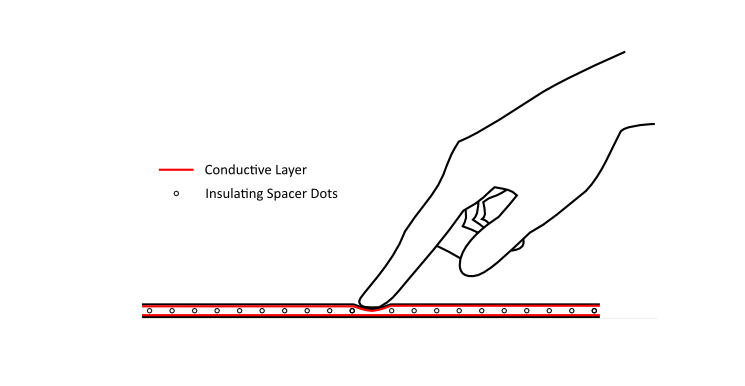}
    \caption{Design of a surface based on resistive technology.}
    \label{fig:tech_resistive}
\end{figure}

Resistive touch screens are typically composed of two layers with uniform resistive coatings of indium tin oxide on their inner sides. The lower layer is typically made of glass or acrylic while the upper layer consists of a flexible hard-coated membrane. The two panels are separated by an insulating layer, usually small invisible spacer dots, so that in idle state the conductive layers are not connected. However when pressed by a arbitrary object the conductive layers connect and a current is established (see figure \ref{fig:tech_resistive}). In order to measure the actual touch position two consecutive electric measurements along the horizontal and vertical axis are performed. Measurements are conducted based on the two following wire configurations.
\paragraph{4-wire configuration}
The 4-wire configuration is the easiest and most common configuration where the inner and outer layer are used to establish the touch position. The resistive coating of the upper layer usually consists of wires running vertically, while the coating of the lower layer consists of wires running horizontally. First a voltage is applied to the upper layer forming a uniform voltage gradient along the wires. When pressed the layers connect and the voltage at the touch point can be measured by the horizontal wires of the lower layer. The horizontal touch position is similarly established by inverting the process and applying a voltage to the lower layer and measuring on the upper layer.

An important drawback of this configuration is it's reliance for touch measurements on consistent electric characteristics of both resistive coatings. However the flexibility of the upper layer will result in microscopic cracks in its coating changing the electric uniformity along the vertical axis. Therefore resistive touch screens exhibit a certain inaccuracy along one axis over time.

\paragraph{5-wire configuration}
The 5-wire configuration solves the shortcoming of diminished accuracy by including both vertical and horizontal wires in the coating of the lower and more stable layer while the upper layer's coating now consists of a single additional wire (therefore the name 5-wire). Similarly to the 4-wire configuration voltage is first applied vertically and then horizontally. The touch position is then determined by the voltage measured using the wire on the upper layer. Applying voltage only to wires on the lower layer ensures consistent and uniform electric characteristics over time. Therefore extensive pressure on the flexible upper layer no more degrades touch detection any more.

\subsection{Capacitive Sensing}
Capacitive sensing relies on the concept of electric capacity occurring between any two conductive elements placed in proximity. The resulting capacity is determined by the dielectric (the layer filling the gap between the conductors), the surface of the conductors and their distance from each other. Its usage for finger touch sensing is enabled thanks to the conductive property of fingers and human skin in general as they contains conductive electrolytes. Capacitive sensing technology is based on two basic principles: self-capacitance and mutual capacitance.

\paragraph{Self-capacitance}
Considering two conductive materials separated by an insulator. If a charge is applied to one of the two materials, an electric field is created on its surface and capacitive coupling occurs resulting in a charge transfer. The amount of charge transferred depends on the distance between the two materials and the insulator.

\paragraph{Mutual capacitance}
Given two conductive materials that form a capacitor as described in \textit{self-capacitance}. Their electric field is mainly concentrated between the conductive materials, however \textit{fringe fields} exist outside this region. Consider such a capacitor and another conductive material placed in its proximity only separated by an insulator. Capacitive coupling occurs due to the \textit{fringe fields} resulting in a charge transfer and a change in electric characteristics of the capacitor.

\subsubsection{Surface capacitive sensing}
\begin{figure}[t]
	\centering
    \includegraphics[width=0.9\textwidth]{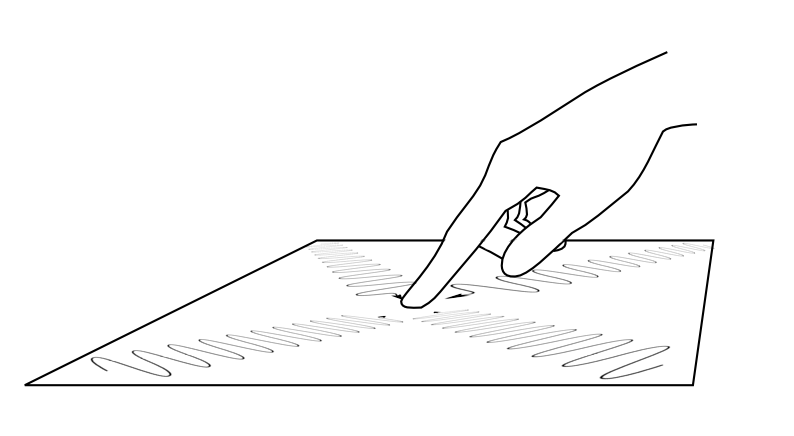}
    \caption{Principle of operation of a surface based on \textit{surface capacitance sensing}.}
    \label{fig:tech_surface_cap}
\end{figure}

Surface capacitance based sensing uses a panel covered by a uniform conductive coating on one of its sides. On all four corners of the conductive coating a charge is applied resulting in a uniform spread of electrons across the surface. Exploiting the principle of self-capacitance a finger approaching the non-coated side of the panel forms a capacitor with the conductive coating at the closest point on the surface. Charge is transferred between the coating and the finger resulting in current on the surface being drawn from the corners (see figure \ref{fig:tech_surface_cap}). Using sensors placed at all four corners, the amount of transferred current is measured. As this amount is inversely proportional to the distance from the touch position, the exact position can be established using bilinear filtering.

Although the technology can be made highly durable by using a durable panel as surface, it suffers from the drawback of being single-touch only. Furthermore its accuracy can be affected by other conductive materials near the touch surface.

\subsubsection{Projected capacitive sensing}
Unlike surface capacitance based sensing, devices based on projected capacitive sensing consist of a grid of spatially separated electrodes patterned on either one or two panels \cite{barrett2010projected}. This grid is usually composed of electrodes in rows and columns and can be operated using the principle of self-capacitance or mutual capacitance. In two-layered devices based on self-capacitance a charge is sequentially applied to all rows and columns and their capacitance is measured. However this defines touch points only as a set of points on the X- and Y-axis without revealing their exact correspondence. This correspondence is obvious for single touch, however is undefined when sensing multiple touch points resulting in so-called ghost points.

A far more common, yet more complex approach is to use electrodes patterned into a single layer measuring electrode intersections instead of rows and columns of electrodes only. Each intersection of electrodes forms a small capacitor when a charge is applied one-by-one to all electrodes in columns. This capacity is measured on each electrode in rows and remains constant as long as no conductive material is placed in its proximity. However given the principle of mutual capacitance a finger approaching the touch surface causes a loss of capacitance on electrode intersections in the touched surface region. Although the grid of electrodes might seem too large on first glance to create accurate touch measurements, one is reminded that the presence and touch of a finger affects a large number of intersections to an extend proportionally inverse to their distance. Given this set of capacity measurements the actual touch position can be accurately determined using bicubic interpolation.

Rekimoto used this technique to create a large scale touchable surface of 80x90cm \cite{rekimoto2002smartskin}. Although he only used an electrode grid of 8x9 cells, hence a cell size of 10x10cm, he achieved a touch accuracy of 1 cm.

With the advent of multi-touch technology in everyday life, projected capacitive sensing is now used on most small-scale devices such as the iPhone \cite{hotelling2010multipoint}.

\subsubsection{MERL DiamondTouch}\label{section:tech_diamondtouch}
DiamondTouch is a touch-sensitive input device developed at Mitsubishi Electric Research Laboratories (MERL) based on capacitive sensing whose capabilities however go beyond the common finger touch detection \cite{dietz2001diamondtouch}. Apart from accurately detecting multiple touch points it reliably identifies for each touch point the corresponding user. Furthermore operation is unaffected by objects placed on or near the device regardless of them being conductive or not.

The device operates using an array of antennas placed below a protective surface acting as an insulator to objects above the device. Each antenna is driven with a uniquely identifiable signal and shielded from neighboring antennas to avoid signal interference. Unlike other devices the electric capacity resulting from conductive objects approaching the surface is not measured inside the device itself but is used to transfer the unique signal of an antenna to this object. Once the user touches the surface a capacitively coupled circuit running from the transmitter (the touched antenna) to a receiver attached to the user's chair is completed. 

Given the presence of current at the receiver a multitude of signals coming from different antennas must be separated by the system. Therefore code-division-multiplexing is used to make the antenna signals mutually orthogonal. Nonetheless a received signal only indicates the antenna being touched, but not exactly where on the antenna. Therefore the antenna size in the proposed design covers only an area of $0.5$x$0.5$mm, an area smaller than the usual finger contact area. Whenever the finger contact area spans several antennas at once an even finer resolution might be achieved by interpolating between the antennas given their relative signal strength.

\subsection{Acoustic Sensing}
Acoustic sensing did only have limited impact in multi-touch sensing technology but exhibits some interesting properties and is therefore included here. Recently acoustic sensing has been used to complement traditional sensing technologies and to provide information on how the surface is being touched\cite{harrison2011tapsense}.

\subsubsection{Location Template Matching}
Location Template Matching (LTM) is based on the principle of Time Reversal. Time Reversal relies on the reciprocity of the wave equation and states that given a recorded acoustic signal the source signal at its original location can be reconstructed by focusing it back in time (by using negative time) \cite{pham2005tangible,fink1993time}. Therefore the recorded acoustic signal itself must be unique as well. Given a single sensor attached to an arbitrary surface, LTM exploits this principle by prerecording a certain number of touches associated with their location. When a user now touches the surface a cross-correlation is applied to the recorded signal to find its best match among the prerecorded ones. Although its main advantage being its aptitude to be used with irregular surfaces (see \cite{crevoisier2008transforming}), touch detection is constraint to a finite number of contact points.

\subsubsection{Time Delay of Arrival}
Whenever the user touches a surface acoustic vibrations are created that propagate along the surface. The Time Delay of Arrival (TDOA) approach measures the incidence of these vibrations at a number of sensor locations. Given the relative delay between distant sensor locations the actual touch location can be inferred (see \cite{pham2005tangible} for more details). The obvious setup would be a flat surface with four sensors attached in rectangular shape, however it could be equally applied to curved surfaces. Nonetheless estimating the propagation of vibrations along a curved surface makes calculations much more computationally demanding.

The main advantage of TDOA is its independence from a specific surface material. Almost all materials such as wood, glass, plastic or metal can be used as long as they transmit acoustic vibrations \cite{crevoisier2008transforming}. On the other hand TDOA is inherently single-touch only and furthermore only detects the incidence of a touch but is ignorant about the objects state afterwards, whether it is gone or still remains on the surfaces.

\subsubsection{Surface Acoustic Wave Absorption}
\begin{figure}[t]
	\centering
    \includegraphics[width=0.8\textwidth]{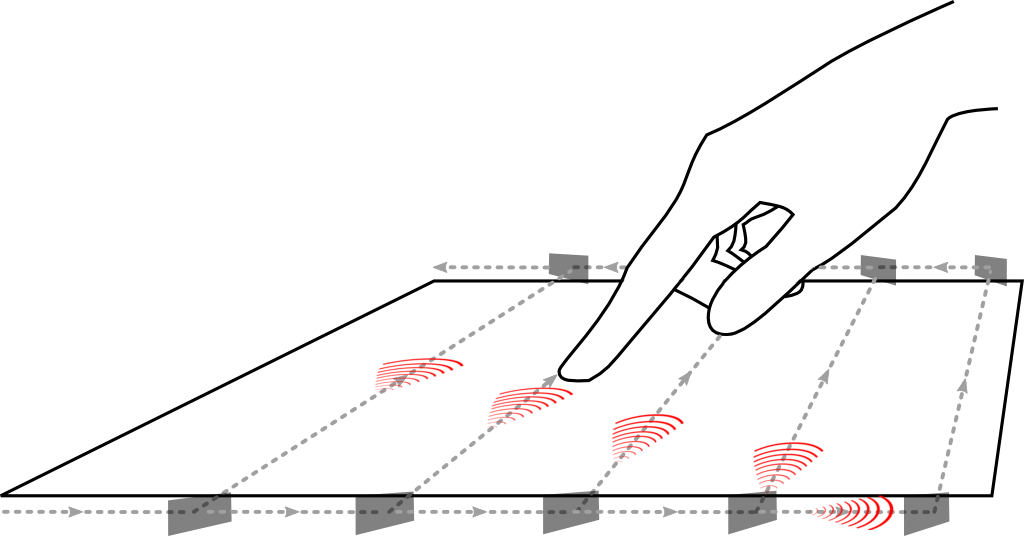}
    \caption{Design of a surface based on \textit{surface acoustic wave absorption}.}
    \label{fig:tech_saw}
\end{figure}

The earliest touch surface based on Surface Acoustic Wave (SAW) absorption ranges back as early as 1972 \cite{johnson1972touch}. The surface panel is equipped on all of its sides with an array of transducers, one array of transducers acting as transmitter while the array opposite acts as receiver. Transmitted surface waves are absorbed by soft objects such as fingers attenuating the measured signal at the receiver. Transducers need to cover the complete contour of the surface to achieve reasonable detection resolution. To reduce the required number of transducers \citeauthor{adler1987economical} proposed the use of an reflective array \cite{adler1987economical}. In this case transducers are placed at the corners of the surface with two of them acting as transmitter and sending a burst of waves along the edges of the surface. These waves traverse an array of partial mirrors each of them reflecting part of the wave vertically inside the surface panel which is then again reflected by a second array of partial mirrors on the opposite side to the receiver (see figure \ref{fig:tech_saw}). An initial pulsed signal therefore results in a continuous response at the receiver due to differing lengths of wave paths. Hence temporal separation determines the response of each single wave path. User interaction with the surface therefore results in a signal attenuation at the corresponding point in time.

As SAW attenuation is proportional to the size of the contact area, one can furthermore infer the amount of pressure applied to a certain contact point allowing to distinguish different types of touch. 

\subsection{Optical Sensing}
Optical sensing has been widely used for touch-sensitive devices because of the richness of data that it provides. Particularly it does not limit the number of touch points and can sense both objects that are in direct contact with the surface and those that only hover above it. However it is exactly the complexity of this data that is sometimes thwarting the intended development of real-time applications.

Camera-based setups using near-infrared light are among the most popular approaches to multi-touch sensing. This is mainly due to the fact that infrared light is invisible to the human eye as well as the increasing availability of cheap cameras that provide high resolutions and perform at frame rates suitable for real-time applications. Furthermore one does not even require a particular infrared camera as digital cameras are sensible to near-infrared light which is generally filtered to avoid noise though. However removing the IR-cut-filter and adding a IR-band-pass-filter in front of the camera can turn any off-the-shelf camera into a fully functional infrared sensing device.

In the following section most of the described approaches rely on cameras for image acquisition hence require a significant throw-distance from the surface. As this might not be viable or desirable for all application scenarios, \textit{intrinsically integrated} devices have emerged recently that directly integrate sensing technology into the surface. Finally an approach based on thermal sensing is described that might not replace the use of near-infrared light for multi-touch sensing but rather complement it given its several desirable sensing properties.

\subsubsection{FTIR}\label{ftir}
The rediscovery of \textit{frustrated total internal reflection} (FTIR) by \citeauthor{han2005low} and its application in multi-touch sensing resulted in increased interest in optical sensing technologies due to its simple, inexpensive and scalable design \cite{han2005low}. FTIR relies on the optical phenomenon of \textit{total internal reflection} that can occur when a ray of light strikes a boundary with a medium of lower refractive index. The amount of light being refracted depends on its angle of incidence with more light being reflected back into its medium of origin the more acute the angle. If the angle of incidence is beyond a \textit{critical} angle all of the light is reflected resulting in \textit{total internal reflection}. However once the initial assumption of a lower refractive index is violated by a different medium, it \textit{frustrates} the \textit{total internal reflection} and light escapes at the boundary.

\begin{figure}[t]
	\centering
    \includegraphics[width=0.9\textwidth]{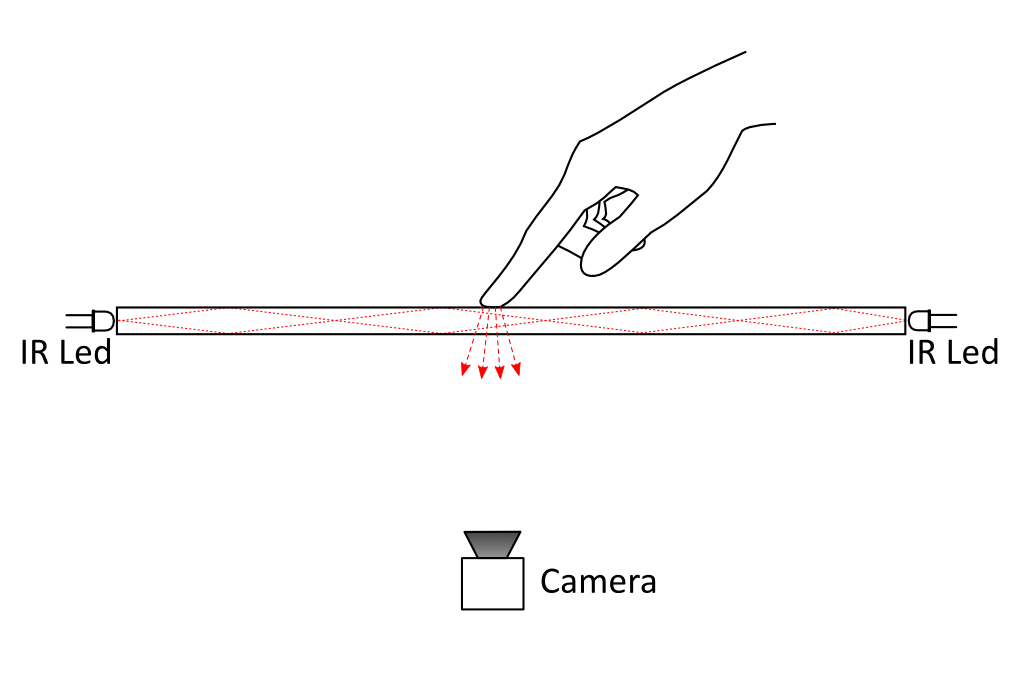}
    \caption{Design of a surface based on \textit{frustrated total internal reflection}.}
    \label{fig:tech_optical_ftir}
\end{figure}

This principle is applied to multi-touch devices using a sheet of acrylic or a similar material with high optical transmittance. Along the edges arrays of infrared LEDs are located that inject light into the panel. Due to the parallel orientation of the LEDs to the panel's surface most infrared light is internally reflected. However when an object with higher refractive index such as a finger touches the surface light escapes from the area of contact. This light is then reflected from the finger and can be captured by an infrared-sensitive camera placed behind the surface (see figure \ref{fig:tech_optical_ftir}).

The advantage of this approach is its ability to only identify objects that actually touch the surface. However it is at the same time its main drawback as contact areas lack any contextual information. Contact areas identical in shape could be caused by the same or completely different objects. Contact areas in proximity could be caused by the same or again completely different objects. This missing contextual information makes identification and classification, such as association of a finger to a hand and ultimately a user, extremely difficult.

As the setup uses a camera to capture the surface a significant amount of space is required behind the surface to accommodate the camera. The larger the surface the further the camera needs to be located behind it which might not be feasible in all application scenarios. This could be overcome using two or more cameras that each only capture part of the surface or by using wide-angle cameras. However the former would require further synchronization and calibration procedures making the setup more complex while the latter would reduce the actual resolution at off-axis points due to barrel distortion.

In \cite{hofer2009flatir} a prototype is described that allows incorporation of FTIR technology in common LC-screens hence enabling very thin form factors. This prototype is in its setup inspired by the work on ThinSight (see page \pageref{tech_optical_integrated}) and uses as well a large array of infrared sensors mounted behind the LC-matrix to capture incoming light. Acrylic fitted with a frame of LEDs is placed in front of the LC-screen. They claim to achieve update rates of at least 200 Hz with a 1mm resolution however make assumptions on touch pressure and finger orientation in order to trigger a touch event.

\subsubsection{Diffuse Illumination}
Diffuse Illumination is a simple design of multi-touch sensing technology requiring solely a diffusing surface, a infrared-sensitive video camera and an infrared illuminator. One distinguishes two types of diffuse illumination:

\begin{description}
\item[Back Illumination] Back illumination describes a setup where both the camera and the illuminator are located behind the surface (see figure \ref{fig:tech_optical_di_back}). Due to the diffusing capabilities of the surface, the infrared light emitted from the illuminator creates a well lit area in front of the surface illuminating any object in its proximity. The video camera subsequently captures these objects which can then be identified using common image processing techniques. 

\begin{figure}[t]
	\centering
    \includegraphics[width=0.9\textwidth]{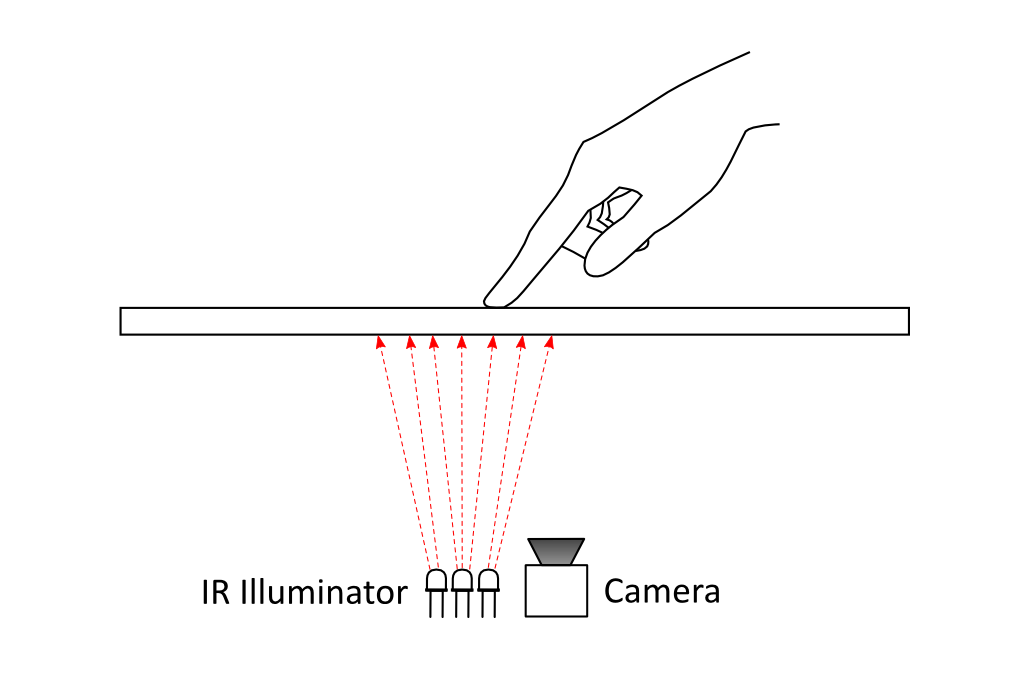}
    \caption{Design of a surface based on \textit{diffuse back illumination}.}
    \label{fig:tech_optical_di_back}
\end{figure}

The inherent simplicity of this approach which can be constructed at low-cost with no required engineering skills makes it a widespread alternative to other more sophisticated technologies. Furthermore this setup can be scaled up to almost any size allowing a variety of different application scenarios. However this simplicity comes with a number of shortcomings. The ability to sense objects near the surface, therefore providing contextual information, actually complicates differentiating touching from hovering the surface. Although objects touching the surface should theoretically reflect much more light to the camera, hence providing a strong criteria for differentiation, yet it is in practice almost impossible to achieve a uniform illumination of the surface. Moreover external infrared light sources and stray sun light might interfere with object detection.

A well known commercial application of this technology is the Microsoft Surface\footnote{see \url{http://www.surface.com}} (Surface 1) which uses in total 4 cameras to capture the surface. Achieving a net resolution of 1280x960 they can detect arbitrary objects placed on the surface based on their shape as well as small marker patches. The Surface 1 has now been replaced by the Surface 2 which follows an intrinsically integrated approach to multi-touch sensing (see page \pageref{tech_optical_integrated}).

\item[Front Illumination]
Unlike back illumination, this approach places the infrared illuminator on the opposite side of the video camera (see figure \ref{fig:tech_optical_di_front}). Hence the closer an object is located to the surface the less light is reflected towards the camera. This approach minimizes the interference of external infrared light with object detection however might suffer from occlusion effects. Another finger, the hand or even the arm might obscure the contact area of a touching finger if placed too close to each other therefore largely reducing the contrast in the resulting camera image. Due to these shortcomings back illumination is often preferred to front illumination.

\begin{figure}[t]
	\centering
    \includegraphics[width=0.9\textwidth]{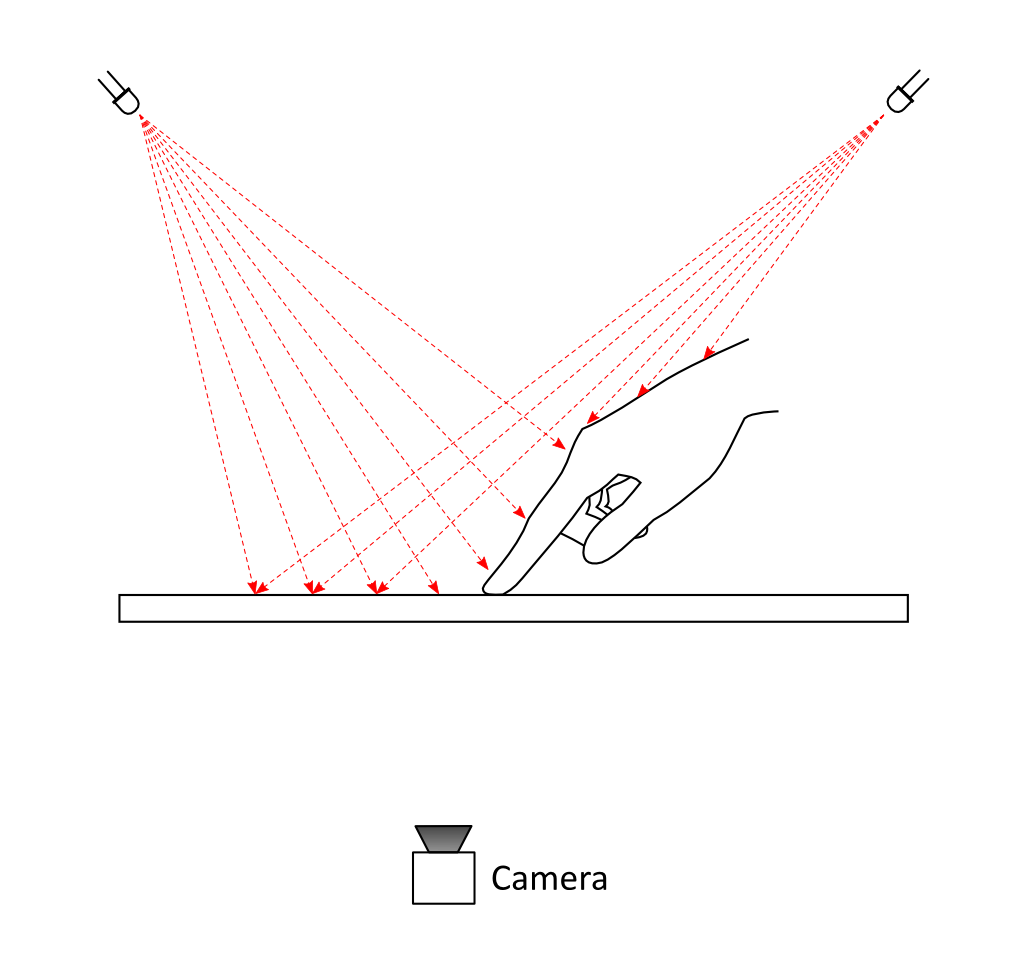}
    \caption{Design of a surface based on \textit{diffuse front illumination}.}
    \label{fig:tech_optical_di_front}
\end{figure}

\end{description}

\subsubsection{Diffused Surface Illumination}
\begin{figure}[t]
	\centering
    \includegraphics[width=0.9\textwidth]{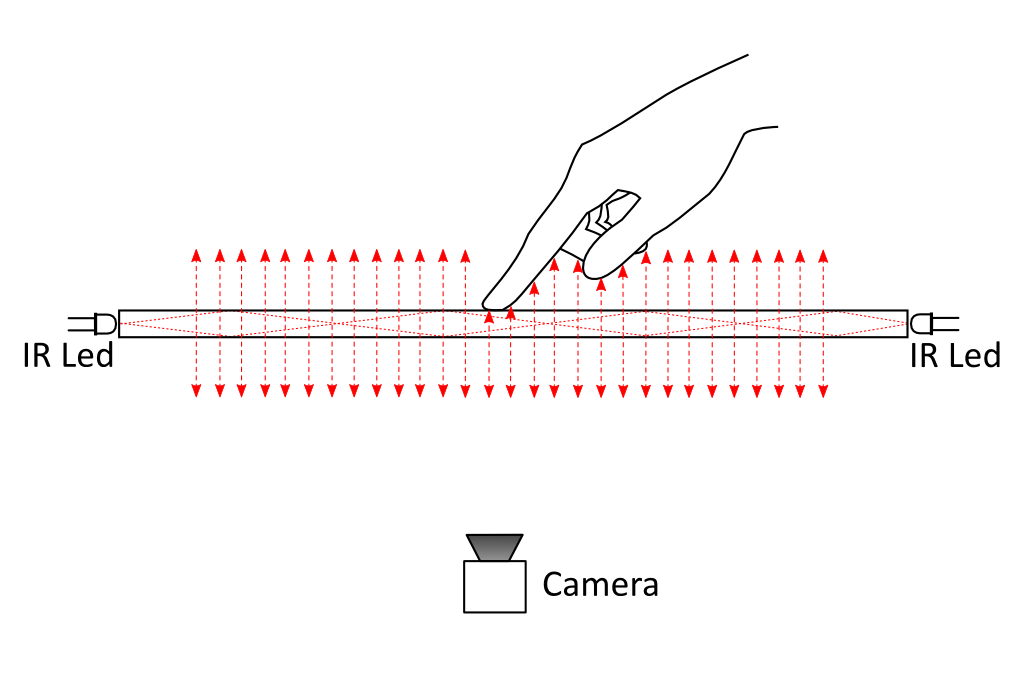}
    \caption{Design of a surface based on \textit{diffused surface illumination}.}
    \label{fig:tech_optical_dsi}
\end{figure}

Diffused Surface Illumination describes a variation of diffused illumination that overcomes the shortcoming of uneven illumination \cite{roth2008dsi}. As for FTIR the setup consists of a sheet of acrylic contained in a frame of infrared LEDs, however this time a special type of acrylic called EndLighten is being used. This acrylic can be thought of as being filled with numerous reflective particles evenly spread across the sheet. When light enters the acrylic, it is reflected by these particles evenly to both sides of the sheet resulting in uniform illumination of objects in proximity (see figure \ref{fig:tech_optical_dsi}). However as light is reflected towards the camera as well the captured image exhibits less contrast compared to the regular approach. Given the reduced contrast the setup is now much more sensitive to external light interference.

\subsubsection{Infrared Light Plane}
The idea behind this approach is to use a setup similar to diffused illumination but to only illuminate the space just above the surface. A finger approaching the surface and therefore entering the light plane reflects light to the camera mounted behind the surface. The light plane can be spanned either by mounting infrared LEDs around the surface area or by placing IR lasers with line lenses at all four corners. To a certain extend this technology might suffer from occlusion effects as objects blocking a light beam reduce the amount of IR light reflected by the occluded object.

The description of a prototype build using this technology on top of a LCD-screen can be found in \cite{motamedi2008hd}.

\subsubsection{Infrared Light Beam Grid}
Similarly to other grid based approaches, this technology spans a grid of infrared light beams across a surface identifying surface contact by measuring occlusion of light beams. Displays of any size can be fitted with that technology by mounting a frame of infrared transmitter and receivers on top of the display. In this frame transmitters and receivers are arranged just above the surface on opposite sides horizontally and vertically. Each light beam is tested for occlusion by sequentially pulsing infrared transmitters and simultaneously measuring incoming light by the receivers on the opposite side. If an object or a finger approaches the surface it obstructs the light path and reveals its presence. Given the horizontal and vertical intersections the touch coordinates and an axis-aligned bounding box of the contact area can be determined. However similarly to other grid approaches this technology might suffer from occlusion effects with multiple, simultaneous touch points. 

\subsubsection{Intrinsically Integrated Sensing}\label{tech_optical_integrated}
In \cite{hodges2007thinsight} the concept of intrinsically integrated sensing is introduced for technologies that integrate sensing capabilities directly into the display and therefore distribute sensing capabilities across the surface. While previous camera-based approaches relied on a perspective projection of objects, intrinsically integrated sensing can be thought of as an orthogonal projection making detection and recognition of objects much more reliable regardless of their position. 

In this context an important finding is that LEDs might not only act as light emitter but under certain circumstances also as light receiver which goes back to F. Mims in 1973 \cite{mims1973led}. Although not being widely adapted for a long time, \cite{dietz2003very} presents first applications and in \cite{hudson2004using} it is introduced as sensing technology of a touch-sensitive device prototype. Han later presents a similar sensor for proximity sensing \cite{han2005led,han2009multi}, while in 2010 \citeauthor{echtler2010led} report to have fitted such a LED array in a 17’’ LCD-screen \cite{echtler2010led}. 

\citeauthor{hodges2007thinsight} describe a prototype of intrinsically integrated sensing called ThinSight \cite{hodges2007thinsight} that is based on an array of retro-reflective optosensors. These components consist of both a light emitter and receiver that are optically shielded from each other. They allow emitting and sensing infrared light simultaneously resulting in a gray level intensity image of reflected infrared light above the surface. Similarly to diffuse illumination they propose two modes of operation named \textit{reflective mode} and \textit{shadow mode} that loosely correspond to \textit{diffuse back illumination} and \textit{diffuse front illumination} respectively. \textit{Reflective mode} relies on the emitters to illuminate the area above the surface tracking bright regions while \textit{shadow mode} relies on ambient infrared light for illumination tracking shadows of objects in proximity. Furthermore ThinSight allows novel interaction techniques with any simultaneous number of devices that allow communication using modulated infrared light such as smartphones or pointing sticks.

The only commercial application of this technique seems to be the most recent version of the Microsoft Surface\footnote{see \url{http://www.surface.com}} (Surface 2 or Samsung SUR40). Its main characteristic is its very slim form factor that they attribute to a technology called PixelSense \cite{surface2011pixelsense,den2011infrared} which is in fact very similar to the described ThinSight above.

\subsubsection{Depth sensing}
The use of two cameras for touch sensing might at first glance seem like an unnecessary overhead due to increased computational demand but several interesting and yet performant applications have been developed. 

\begin{description}
\item[TouchLight]
TouchLight \cite{wilson2004touchlight} is an interactive wall prototype that uses a stereoscopic setup combined with a back-projected, transparent surface (DNP HoloScreen). While previous setups employed diffuse surfaces that explicitly limit the camera to sense solely objects in proximity of the surface, the transparent surface does not place any restrictions on sensing. Moreover without diffusion object sensing is ultimately more accurate enabling interaction techniques like scanning documents placed on the surface or identifying the current user by means of face recognition. Instead of focusing just on pointing interaction one is more interested in creating a richer representation of objects on the surface describing their size and shape using a \textit{touch image}.

Given the binocular disparity resulting from the different camera viewpoints objects on the projection surface can be accurately extracted. Using simple image processing steps such as applying transformations based on homography matrices and computing edge filters the \textit{touch image} for the projection surface can be computed (see \cite{wilson2004touchlight} for a detailed description). Furthermore the interaction plane is not restricted to lie exactly on the projection surface but could be represented by any plane in space enabling what they call \textit{Minority Report} interfaces.

\item[PlayAnywhere]
PlayAnywhere \cite{wilson2005playanywhere} is a compact and portable prototype consisting of a projector, a camera and an infrared illuminant that allows to transform any flat surface into an interactive area. Sensing of touch is achieved by measuring the disparity between a fingertip and its shadow on the surface. As the illuminant is placed off-axis from the camera the position of a finger hovering the surface differs from its corresponding shadow position while on touch these positions coincide. Furthermore as the distance between illuminant and camera is fixed a three-dimensional position of the fingertip might be computed. However this technique may suffer from occlusion and in the current implementation applies certain restrictions such as supporting only one finger per hand. 

Additionally another approach is described to allow basic interaction techniques such as panning, rotation and scaling of the viewport without actually identifying a users hand or finger. Given the direct sight of the camera the optical flow field can be extracted from a sequence of camera images from which the corresponding transform matrix consisting of scaling, rotation and translation can be derived.

\item[Kinect]
Since the Microsoft Kinect, a low-cost yet reliable off-the-shelf alternative to much more expensive depth-sensing systems, has become widely available much research has focused on this system. In \cite{wilson2010using} a touch sensing system using a Kinect mounted above the surface is described. Although touch sensing might seem straightforward with such a system it turned out to be actually impossible without previous assumptions. Fingers differ in thickness between each other and between people making it hard to set a fixed distance threshold indicating whether the finger touched the surface. Furthermore if the finger is oriented at a steep angle with respect to the surface the finger might obstruct the direct line of sight between camera and its fingertip.

Assuming a direct line of sight between the camera and fingertips pixels are classified as touching if they are within a predefined distance from the prerecorded background. Although this technique does not match the sensing accuracy of previously described approaches it easily allows sensing touch on arbitrarily shaped surfaces. Furthermore the shape information of objects above the surface enables grouping touch points to hands and ultimately a user. 
\end{description}

\subsubsection{Thermal Sensing}
Thermal sensing has so far not been widely explored in the field of touch-sensing devices (only EnhancedDesk2 in \cite{koike2001integrating} and ThermoTablet in \cite{iwai2005heat} are known to the author). Unlike previous approaches that operate in the near-infrared spectrum, thermal sensing measures the amount of far-infrared light emitted by objects that is related to their temperature. As presented in \cite{larson2011heatwave} thermal sensing has several desirable properties that can complement existing approaches of touch-sensing. An obvious advantage is its independence of lighting conditions as it works equally well with sun light as with indoor lighting or darkness. Therefore it can be used to reliably segment hands from the background. A further very interesting property is the heat transfer that takes place when a finger touches a surface. Not only is the amount of heat at the contact area a measure of touch pressure but also remains at its position for a certain time after the finger has been lifted. When dragging a finger across the surface heat traces are left behind indicating a finger's movement trajectories therefore simplifying the differentiation between touching and hovering and the classification of even complex gestures.

The described prototype confirms the robustness of the the previously described features and reveals some more advantages. Most notably a user's hand seems to have a unique thermal signature that remains constant over the period of interaction. Therefore users can be identified again if they have left the area visible to the camera. Furthermore this thermal signature applies as well to objects of different materials that are placed on the surface allowing the differentiation of materials.

\section{Processing Pipeline}
This section will detail the required processing steps to reveal contact points on a surface based on a diffuse back-illumination approach. This processing pipeline will be considered the de facto standard and serves as a basis for comparison with the novel processing pipeline presented in the next chapter.

Although the diffuse back-illumination approach is widely used and documented in research as well as in the wider multi-touch community, surprisingly little can be found about the actual image processing steps in the literature. While a simple set of image processing operation appears to be the default approach, it requires certain assumptions that may not be valid in all circumstances. The following processing pipeline details these essential steps which have been extracted from prototypes described in literature and approaches used in the open-source multi-touch community, namely the Community Core Vision framework\footnote{see \url{http://ccv.nuigroup.com/}}.

\subsection{Definitions}\label{basic_definitions}
In order to establish a common understanding of the image processing operations in this as well as in the following chapter, a number of basic definitions are being presented here. These definitions mainly cover notation and simple concepts related to images and image areas.

\begin{description}
\item[Blob] \hfill\\
A \textit{blob} describes a region of an image that is being considered to originate from an object that is currently touching the surface. However a \textit{blob} does not imply any sort of semantic with respect to the object being in contact with the surface. The terms \textit{blob} and \textit{contact point} are both being used interchangeably throughout the thesis.

\item[Image] \hfill\\ 
An Image $I$ is a mapping of two-dimensional coordinates to an intensity value: $I: C_I \rightarrow V$ with $C_I\subset \mathbb{Z}^2$. Intensity values are generally encoded as 8-bit values, therefore $V$ is defined as $\{0,\dots,255\}$. If  $C_I = A_I\times B_I$ the size of an image $I$ is defined as $|A_I|*|B_I|$.

\item[Pixel] \hfill\\ 
A pixel $p = (x,y)$ is the smallest entity of an image $I$: $(x,y) \in C_I$. It is represented by its coordinates $(x, y)$ and its intensity value $I(x, y)$.

\item[Neighborhood] \hfill\\ 
For each pixel $p$ in $I$ a neighborhood relation $N(p)$ is defined representing its neighboring pixel. Most common in image processing are 4- and 8-neighborhood relations.
\begin{description}
\item[4-Neighborhood] \hfill\\ 
Given an image $I$ with pixels arranged in an orthogonal grid, then the 4-neighborhood of a pixel $p$ is defined as follows:
\begin{equation}
N(p)=N(x,y) = \left\{ (m,n) \mid (\mid m - x\mid + \mid n-y\mid = 1) \wedge (m,n) \in C_I \right\}
\end{equation}
\item[8-Neighborhood] \hfill\\ 
Given an image $I$ with pixels arranged in an orthogonal grid, then the 8-neighborhood of a pixel $p$ is defined as follows:
\begin{equation}
N(p) = N(x,y) = \left\{ (m,n) \mid (\max(\mid m - x\mid, \mid n-y\mid) = 1) \wedge (m,n) \in C_I \right\}
\end{equation}
\end{description}
Other neighborhood relations exist, for instance in hexagonal grids, however these are outside the scope of this thesis.

\item[Path]\hfill\\
A path between two pixels $a$ and $b$ in an image $I$ exists if and only if there exists a set of pixels $P_{a,b} = \{a, p_1, \dots, p_n, b\} \subseteq C_I$ such that
\begin{equation}
p_1 \in N(a) \wedge b \in N(p_n) \wedge \forall i\in\{1,\dots,n-1\}\mid p_{i+1} \in N(p_i)
\end{equation}

\item[Region]\hfill\\
A region $R$ is a connected set of pixels from an image $I$ such that:
\begin{equation}
R \subseteq C_I \wedge \left(\forall a,b\in R\,\exists P_{a,b}\mid P_{a,b}\subseteq R\right)
\end{equation}

\item[Inner Region Boundary]\hfill\\
The inner boundary $\omega_R$ of a region $R$ is the set of pixels inside the region connected to at least one pixel outside the region:
\begin{equation}
\omega_R = \left\{ p \mid p\in R \,\wedge ((N(p) \cap C_I \setminus R) \neq \varnothing) \right\}
\end{equation}

\item[Outer Region Boundary]\hfill\\
The outer boundary $\Omega_R$ of a region $R$ is the set of pixels outside the region connected to at least one pixel inside the region:
\begin{equation}
\Omega_R = \left\{ p \mid p\notin R \,\wedge (N(p)\cap R \neq \varnothing) \right\}
\end{equation}
\end{description}

\subsection{Blob Detection}\label{traditional_processing_pipeline}
The first part of the processing pipeline, namely \textit{blob detection}, reveals all objects that are in contact with the multi-touch surface. Some approaches try to limit the number of contact points by removing all image information originating from objects that are either too large or too small to be considered an intentional touch from a user's finger. Nonetheless the basic steps of this part of the pipeline are as follows:
\begin{itemize}
\item Background Subtraction
\item Image Filtering
\item Thresholding
\item Connected Component Analysis
\end{itemize}

\subsubsection{Background Subtraction}\label{background_subtraction}
As for all setups based on infrared illumination, the detection accuracy can be negatively affected by ambient infrared light such as stray sun light. Hence the first step is to remove interfering light from the image using \textit{background subtraction}. This operation subtracts a previously defined reference image pixel by pixel from a newly acquired camera image.

The reference image can be acquired in two ways. Firstly a prerecorded image of the background could be used as a static reference image. Secondly processing could initiate with a prerecorded image and then gradually adjust the reference image to changes in the background. The former approach is suitable to setups where the ambient illumination can be controlled which generally applies to indoor situations. While the latter approach can reduce the effect of background changes it results in an additional computational overhead. Furthermore it requires precise timing of when to update the reference image. Otherwise objects present on or near the surface might be included in the reference image resulting in erroneous detection results.

A reference image is usually computed using the arithmetical mean of a large number of camera frames. A static reference image $R_{static}$ is defined as
\begin{equation}
R_{static}(x,y)=\frac{\sum_i^n{I_i(x,y)}}{n}
\end{equation}
with $I_1,\dots,I_n$ being the prerecorded background images.

A dynamic reference frame $R_{dynamic}^i$ at time $i$ using a running average is defined as
\begin{equation}
\begin{array}{rcl}
R_{dynamic}^i(x,y)&=&(1 - \alpha) * R_{dynamic}^{i-1}(x,y) + \alpha * I^i(x,y)\\
R_{dynamic}^0(x,y)&=&R_{static}(x,y)
\end{array}
\end{equation}
with $I^i$ and $\alpha$ being the camera image at time $i$ and the \textit{learning rate} respectively. The \textit{learning rate} defines the speed in which the reference image adapts to changes in the background.

\subsubsection{Image Filtering}
This step is essential to facilitate the following blob detection by removing noise from the image. Generally everything that is not related to an intentional contact with the surface is considered as noise. Using this definition different types of noise can be distinguished:

\begin{description}
\item[Camera Noise] Digital cameras inherently suffer from two types of image noise known as fixed pattern noise and random noise that are introduced during the image acquisition process. These result in slight variations of pixel intensities and mainly affect high frequencies in the image. Furthermore, especially in low-end cameras, compression artifacts are introduced into the output resulting in a certain "blockiness" of the image.
\item[Debris on the surface] Considering a tabletop surface as its name suggests as a table, the user might place random objects required for the task onto the surface. As these might range from small objects like pens to larger ones like paper, these must not result in erroneously detected touch inputs.
\item[Palm/Wrist/Arm] During user interaction surface contact of any body part other than fingertips is generally considered unintentional.
\end{description}

Most multi-touch prototypes solely focus on reducing the effect of camera noise as the other two are not part of the expected usage scenario. Therefore it is up to the user to not violate these assumptions to ensure a proper functioning of the device. However long periods of interaction on large tabletops usually result in arm fatigue largely increasing the likelihood of unintentional touches. The effect of camera noise is reduced by applying a low-pass filter that eliminates high frequencies.

Some image processing pipelines are designed to remove the latter two types of noise as well (see \cite{wolfe2010seeing}). This is achieved by defining a range of expected input dimensions and filtering objects larger or smaller than this range from the input image. As multi-touch tabletops are generally used with fingers or a stylus as input modality this can be considered a valid assumption. However this approach is incompatible with the use of markers located on the bottom side of objects placed on the surface as these would be removed during filtering. 

The above mentioned low-pass filtering step in image processing is equal to blurring of the image. Simple blurring is achieved by applying a box filter that replaces a pixel's value with the average of the area around it. A more sophisticated form is Gaussian blur where pixel intensities are weighted according to the Gaussian. Blurring of an image $I$ is usually computed using convolution (denoted as $\ast$)
\begin{equation}
I_B(x,y)=I\ast K=\sum_{i=-\lfloor k/2\rfloor}^{\lceil k/2 \rceil}\sum_{j=-\lfloor l/2\rfloor}^{\lceil l/2\rceil}{I(x+i,y+j)\cdot K(\lfloor k/2\rfloor + i,\lfloor l/2\rfloor + j)}
\end{equation}
with $K$ being the convolution kernel of size $k\times l$ having its origin located in its center. The size of the kernel defined according to the box filter or the Gaussian determines the frequency cut-off resulting from the convolution.

An essential property of convolution kernels is \textit{separability}. This property refers to the dimensionality of a kernel and allows the representation of a higher-dimensional kernel using a set of lower-dimensional kernels therefore highly reducing its computational complexity. Consider a two-dimensional kernel. If the kernel is separable, the result of its convolution with an image is equal to the result of sequential convolutions using two one-dimensional kernels. Therefore 
\begin{equation}
K = V \ast H
\end{equation}
with $V$ and $H$ being one-dimensional kernels representing the vertical and horizontal projections respectively. Both the box-filter and the Gaussian filter are separable and can therefore be efficiently computed in image processing.

In order to filter objects outside a size range usually a band-pass filter is used. A band-pass filter can be thought of as the difference of two low-pass filters while their cut-off frequencies define the minimum and maximum size of the target object respectively. A band-pass filter based on two box-filters or two Gaussian filters is called \textit{Mid-Box} or \textit{Mid-Gauss} respectively. In \cite{wolfe2010seeing} a \textit{Dual Quad} filter is proposed as an efficient approximation of the \textit{Mid-Gauss} filter. In their evaluation of different filters within a multi-touch processing pipeline, they found the \textit{Mid-Gauss} and \textit{Dual Quad} to both achieve the highest accuracy while the \textit{Dual Quad} significantly outperforms the \textit{Mid-Gauss}.

\subsubsection{Thresholding}\label{thresholding}
Following the previous background removal and filtering steps, the guiding assumption of this step is that now the intensity of every pixel relates to the distance of the displayed object to the surface. Therefore in this step all those pixels belonging to objects that actually touch the surface are being identified using thresholding.

Thresholding is a simple and widely used image classification operation that assigns each pixel based on its intensity to a class from a predefined set. In image processing binary thresholding is the most common thresholding type used to differentiate between foreground and background. Foreground usually refers to objects of interest while everything else is considered as background. 

Thresholding of an image $I$ is defined as
\begin{equation} \label{equ_thresholding}
T(x,y) = \begin{cases}
1\quad\mbox{if }I(x,y) \geq T  \\
0\quad\mbox{otherwise} 
\end{cases}
\end{equation}
with $T$ being the threshold. 

As illustrated by equation \ref{equ_thresholding}, the result of this operation is solely dependent on $T$ which is why the appropriate choice of $T$ is crucial. One usually differentiates between fixed and adaptive thresholding techniques based on the threshold selection process. In the former case $T$ is chosen manually based on experience or another heuristic that does not rely on knowledge of the actual image content. The latter however uses information extracted from the image content to infer an appropriate threshold value. One of the most well known examples of this selection process is Otsu's method that classifies image pixels in two distinct classes such that the intra-class variance is minimized \cite{otsu1975threshold}.

However the accuracy of the thresholding operation might suffer from noise and non-uniform illumination. The variation of pixel intensities introduced by noise results in misclassification of pixels with intensity close to the threshold value while non-uniform illumination prevents the accurate classification of all pixels considered as foreground. These shortcomings stem from the application of a single threshold value for all pixels regardless of local image characteristics. Therefore one has to further differentiate between global and local thresholding methods. Usually globally operating methods are preferred thanks to their superior performance in real-time systems however local adaptive thresholding has been successfully employed in the reacTIVision system \cite{bencina2005improved}. A vast number of different thresholding techniques exists that are outside the scope of this thesis. The interested reader is referred to \cite{sezgin2004survey} which provides a comprehensive survey of existing techniques.

Returning to the initial assumption that pixel intensities directly relate to object distance two important limiting remarks need to be made.
\begin{description}
\item[Uniform illumination is very hard to achieve in practice] \hfill\\
Although being fundamentally related to the above assumption in practice it proves to be very hard to actually achieve an uniform illumination of large surfaces. To circumvent this problem the number of infrared illuminators is usually kept low however even a single illuminator still spreads light unevenly off axis.
\item[Objects reflect light differently]\hfill\\
Considering the fingers of a hand, significant differences in the amount of light reflected upon touch can be observed. The amount of reflected light depends on the size of the contact area and the pressure applied by the fingertip hereby for instance favoring the index finger compared to the little finger.
\end{description}

Therefore the choice of an appropriate threshold is a challenging task requiring significant manual calibration efforts. Furthermore on large-scale surfaces non-uniform illumination becomes a serious issue making the distinction between hover and actual touch of an object greatly more difficult.

\subsubsection{Connected Component Analysis}\label{section_connected_components}
Up to this step pixels have only been considered individually without considering the possible connectivity between them. As objects span more than a single pixel, a set of neighboring pixels can be grouped assuming that all those pixels represent the same object. Grouping is performed using a neighborhood relation, usually 4-neighborhood, and a homogeneity criterion that must be fulfilled by all pixels included in the set.

\begin{description}
\item[Connected Component]\hfill\\
A region $R$ in an image $I$ represents a connected component $C$ if and only if
\begin{equation}
C = \{p\mid p\in R \wedge H(p, I(x_p, y_p)) = 1\}
\end{equation}
with $H$ being a homogeneity criterion function. $H$ returns $1$ if the criterion holds for a pixel or $0$ otherwise.
\end{description}

As the previous step already performed a homogenization of image pixels by performing binary classification of pixels as either foreground or background, the homogeneity criterion function is simply equivalent to the result of $T(x,y)$.

A naive approach to finding connected components in an image would consist of scanning image pixels row by row and running a region-growing algorithm once a pixel has been found for which $H(p, I(x_p, y_p)) = 1$. However due to the unrestricted nature of the region-growing algorithm a large number of unnecessary comparisons and boundary checks need to be performed. More sophisticated approaches exist such as the sequential algorithm presented by \citeauthor{horn1986robot} which will be described in section \ref{section_connected_components_horn}.

The identified connected components are generally considered valid contact areas resulting from user interaction. However simple plausibility checks comparing their size with predefined minimum and maximum values are still performed in order to discard unreasonably small or large contacts. The remaining components $C_i$ are then reduced to a single contact point, usually their center of mass $c_i$

\begin{equation}
c_i=\left(\frac{1}{\mid C_i\mid}\cdot{\sum\limits_{(x,y)\in C_i}{x}}, \frac{1}{\mid C_i\mid}\cdot{\sum\limits_{(x,y)\in C_i}{y}}\right)
\end{equation}

\subsection{Blob Tracking}\label{section:default_pipline:tracking}
Complementary to blob detection that runs on a per frame basis tracking now aims to establish intra-frame relationships between the detected contact points. Given the frame rates of usually 30 and up to 60 frames per second a simple nearest neighbor matching between the sets of contact points of consecutive frames is generally sufficient. However in case of unsteady motion or multiple contact points in proximity the application of a motion model might be required. Such a model could then be used to predict the current position of a contact point hereby improving the performance of the subsequent nearest neighbor matching. This approach using a Kalman filter has been described by \citeauthor{oka2002real} in \cite{oka2002real} and is outlined in the following section.

Hence \textit{blob tracking} comprises the following two processing steps:
\begin{itemize}
\item Given the position of a blob in previous frames predict its estimated position in the current frame using a motion model, e.g. the Kalman filter.
\item Using the predicted positions of previous blobs in the current frame, establish relationships between blobs using nearest neighbor matching.
\end{itemize}

\subsubsection{Kalman Filter}
The Kalman filter allows reliable predictions of process states at arbitrary points in time both in the past and the future through minimization of the average quadratic error. Furthermore a feedback loop is used that considers the difference between the predicted and measured state at the current point in time in future predictions. Basically the Kalman filter distinguishes \textit{a priori} and \textit{a posteriori} predictions. The former describe state predictions by one time step into the future based on the current state. The latter include both the a priori prediction and the actual measurement hereby taking into consideration the prediction error of the \textit{a priori} prediction.

According to \cite{welch1995introduction} two groups of equations can be distinguished: \textit{time update} and \textit{measurement update} equations. The former serve the prediction of states while the latter take into account the actual measurements. Given these two groups of equations the following algorithmic design is described:

The prediction starts with the calculation of the state at time $k+1$:
\begin{equation}
 \hat{x}^{-}_k = A \cdot\hat{x}_{k-1} + B \cdot u_{k-1}
\end{equation}
with $\hat{x}^{-}_k$ and $\hat{x}_k$ being the \textit{a priori} and \textit{a posteriori} states at time $k$ respectively. $A$ denominates the state transition matrix while $u_{k}$ and $B$ represent deterministic influences onto the system and its dynamics between consecutive time steps.

Additionally to the state the error covariance is predicted using the following formula:
\begin{equation}
 P^{-}_k = AP_{k-1}A^T+Q
\end{equation}
with $P^{-}_k$ and $P_k$ describing the \textit{a priori} and \textit{a posteriori} error covariance respectively. Q represents the \textit{process noise covariance}.

In order to calculate the a posteriori values within this model, the following \textit{measurement update} equations are used. First the \textit{Kalman gain} is calculated:
\begin{equation}
 K_k = P^{-}_k\cdot H^T(HP^{-}_kH^T+R)^{-1}
\end{equation}
$H$ corresponds to the observation matrix and $R$ to the \textit{measurement noise covariance}.

During the next step the \textit{a posteriori} state is determined using the following equation:
\begin{equation}
 \hat{x}_k = \hat{x}^{-}_{k-1} + K_k(z_k - H\hat{x}^{-}_{k-1})
\end{equation}
$z_k$ represents the state measured at time $k$.

Finally the \textit{a posteriori} error covariance is calculated:
\begin{equation}
 P_k=(I-K_kH)P^{-}_k
\end{equation}

For a more detailed overview on the Kalman filter, the extended Kalman filter and the initial value problem the interested reader is referred to \cite{welch1995introduction}.

The Kalman filter is being applied in a wide range of application scenarios and can be easily implemented due to its recursive nature. However the choice of appropriate initial values and the computational complexity of its calculations represent important disadvantages especially in the case of real-time applications.

\subsubsection{Nearest Neighbor Search}
Nearest neighbor search is used to establish the relationship between contact points in successive frames. Be $P_i = {p_{1,i}, \dots, p_{m,i}}$ and $P_{i+1}={p_{1,i+1}, \dots, p_{n,i+1}}$ the contact point sets at frame $i$ and $i+1$ respectively. As nearest neighbor matching can be represented as an optimization problem, the optimal solution corresponds to the set $S \subset P_i \times P_{i+1}$ that minimizes the following equation:
\begin{equation}
\epsilon=\sum\limits_{(p_i, p_{i+1})\in S}{\parallel p_{i+1} - p_i \parallel_2}
\end{equation}
This problem can be solved using a greedy approach that works as follows:
\begin{enumerate}
\item  Be $\Omega = P_i \times P_{i+1}$ the set of all possible combinations of $P_i$ and $P_{i+1}$. $S = {\O}$.
\item \label{nn_start} Find $(p_i,p_{i+1})\in \Omega$ such that $$\forall (q_i, q_{i+1})\quad \mid\quad \parallel p_{i+1} - p_i\parallel_2 \leq \parallel q_{i+1} - q_i\parallel_2$$
\item Add $(p_i, p_{i+1})$ to the set of solutions $$S = S \cup (p_i,p_{i+1}) $$
\item Remove all combinations that include either $p_i$ or $p_{i+1}$ from $\Omega$: $$\Omega = \Omega \backslash \{\left(p_i,q\right) \mid q \in P_{i+1}\} \backslash \{\left(q,p_{i + 1}\right) \mid q \in P_{i}\}$$
\item If $\Omega\neq {\O}$ goto \ref{nn_start}
\end{enumerate}
However due to the polynomial complexity\footnote{The complexity can be shown to be bound by $\mathcal{O}(n^2\cdot\log{n})$ assuming $m=n$ without loss of generality. First the computation of the distance matrix of all possible combinations of points from $P_i$ and $P_{i+1}$ requires $n^2$ calculations resulting in complexity of $\mathcal{O}(n^2)$. As this matrix is symmetric this number can be reduced to $\frac{n^2}{2}$ which does not affect the computational analysis though. In the next step these distances are sorted which can be achieved in $\mathcal{O}(n^2\cdot\log{n})$. Finally iterating all distances takes again $\frac{n^2}{2}$ operations. Hence the upper bound $\mathcal{O}(n^2\cdot\log{n})$.} of the algorithm search can be costly for large $n$.

%% file: Content/Chapter_2.tex
\chapter{Multi-Touch System}\label{chapter:pipeline}
At first glance, one might be wondering as to why one would envisage to develop a novel processing pipeline for optical multi-touch tabletops. In the last years many new devices have been presented that all rely on the same simple steps to process user input acquired by the camera. So there doesn't seem to be much wrong with that approach. However as \citeauthor{benko2010imprecision} point out current multi-touch prototypes exhibit shortcomings that severely impact usability and hence result in increased user frustration \cite{benko2010imprecision}. They argue that for multi-touch systems every touch counts and hence user interaction is easily affected by accidental inputs (also see \cite{ryall2006experiences} for more detail). In these circumstances users however are only confronted with the resulting unintended application behavior and are left guessing on how they caused that erroneous behavior. Since they will be asking themselves what they have done wrong this is obviously a frustrating user experience and reduces acceptance of the technology. Most commonly these accidental inputs originate from objects placed on the table or parts of the user's arm touching the tabletop during interaction. Furthermore on interactive tabletops users instinctively tend to lean forward and rest their arms on the table as if they were sitting or standing around a regular table hereby increasing the probability of such accidental touches \cite{ha2006direct}.

The problem however is not the user performing in unexpected ways but the general assumption that every touch is equal. Given that assumption unintentional touches have a too high potential to interrupt user interaction while multi-user interaction is made quasi impossible. Furthermore with the continuous growth of interactive tabletops multi-user interaction will be the de facto standard and hence the aforementioned assumption can no longer be considered valid. Nonetheless some approaches exist to tackle that issue as will be described in section \ref{section:hand_distinction} but they all simply consist of a processing layer added on top of the processing pipeline. However we are highly convinced that a processing pipeline that integrates both detection of user input as well as hierarchical clustering of these will in the end be a much better and consistent solution to that problem.

Therefore a multi-touch tabletop prototype based on diffused illumination will be presented here that acts as proof-of-concept of a novel processing pipeline. At the heart of the pipeline is an algorithm called Maximally Stable Extremal Regions (MSER) which has been widely used to extract distinguished regions in an image that exhibit certain unique properties. While these properties will be used to detect contact points the algorithm design furthermore allows the hierarchical structuring of these regions that then will be exploited to infer a clustering of contact points. The resulting clusters are subsequently enriched with additional properties as to whether these contacts originate from a left or a right hand and map detected contact points to the fingers of the hand.

First the design of the tabletop prototype will be presented. Then the major part of this chapter will be dedicated to the elaboration of the proposed processing pipeline while in the end details of the implementation will be outlined.

\begin{figure}[t]
	\centering
    \includegraphics[width=0.6\textwidth]{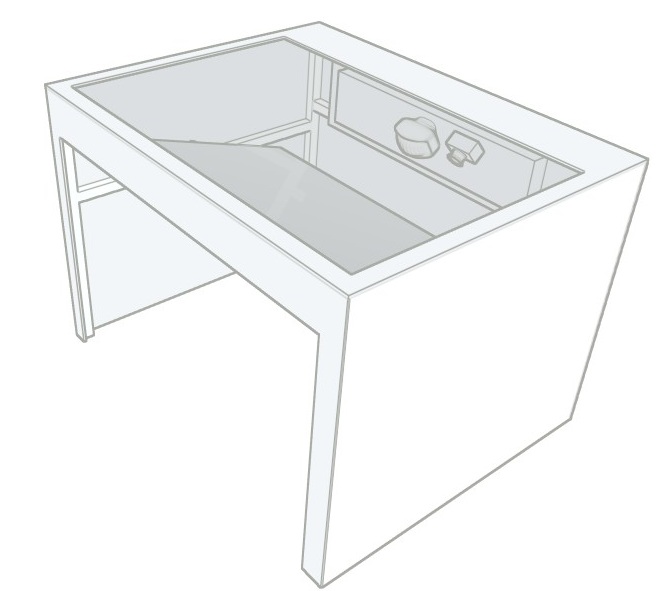}
    \caption{Rendering of the multi-touch prototype.}
    \label{fig:prototype}
\end{figure}

\section{Multi-Touch Tabletop Prototype}
The proposed prototype is based on a setup previously developed by the Virtual Reality Systems Group at the Bauhaus University Weimar. This setup consisted of a DLP projector for image display and a diffuse front-illumination approach for multi-touch sensing. However the latter approach had two significant shortcomings:
\begin{itemize}
\item Due to the large display size several infrared illuminators would be required around the table in order to reduce shadow effects.
\item As the infrared illuminators are external to the tabletop they increase the setup complexity and thwart the desired compactness of the prototype.
\end{itemize}

Therefore a diffuse back-illumination approach was chosen for the new prototype hence integrating both image display and multi-touch sensing into a single device. The setup design is illustrated in figures \ref{fig:prototype} and \ref{fig:prototype_side}. The different components of the prototype are detailed below:

\begin{figure}[t]
	\centering
	\includegraphics[width=\textwidth]{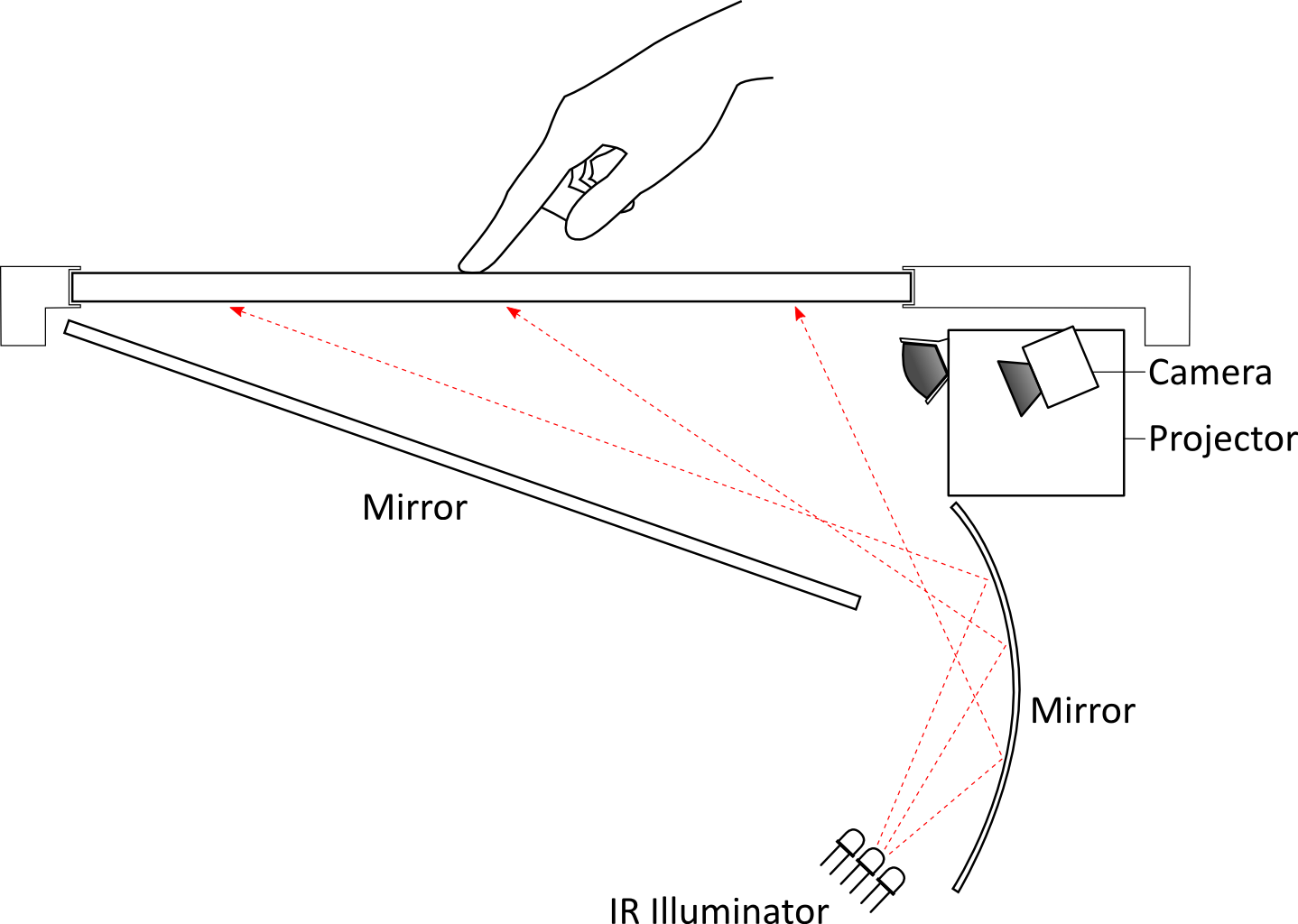}
    \caption{Illustration of the prototype setup.}
    \label{fig:prototype_side}
\end{figure}

\paragraph{Display}
The employed projector is from a dismantled Samsung DLP HL-67A750 67", a high-definition TV capable of displaying active stereo 3D images. Usually projector-based prototypes require a significant amount of space behind the display surface for the projector to cover large surfaces. However thanks to the short throw-distance of this projector a very compact design could be achieved.

\paragraph{Camera}
For image acquisition a Point Grey FireFly MV IEEE 1394a fitted with a fisheye lens is used. It is highly sensitive to the near-IR spectrum while achieving sufficiently fast frame rates (752x480 at 60fps and 320x240 at up to 122fps). As the camera is also sensitive to visible light it had to be fitted with a band-pass filter that blocks light outside the near-IR spectrum. Although many commercial filters are available a piece of developed photo film has been found to work best.

\paragraph{Infrared Illumination}
As the mirror used in the optical path of the projector covers a significant amount of space just below the tabletop surface uniform infrared illumination subsequently proved to be challenging. Moreover the illuminators had not to be directly visible to the camera as overexposure to infrared light might negatively affect sensing of neighboring areas. Therefore possible locations for infrared illuminators were restricted to either beside the camera or below the large mirror. However the first idea had to be dropped due to the large form factor and the limited viewing angle of the infrared illuminators at our disposition during setup design. Hence we opted for the latter approach using an array of infrared illuminators. The employed illuminators (IR SW 77) each consist of 77 LEDs emitting infrared light of 850nm wavelength. In order to spread the light across the whole surface a curved mirror was added to the light path as illustrated in figure \ref{fig:prototype_side}.

\section{Processing Pipeline}
Considering the commonly used processing pipeline as described in section \ref{traditional_processing_pipeline} an efficient method for blob detection exists given the assumption of uniform illumination. This approach is optimized to detect click and drag gestures of finger-like objects on the surface hence supporting the point-and-click metaphor of graphic user interfaces known from desktop computers. However if one tries to go beyond this metaphor and use other properties for user interaction such as which contact points belong to the same hand one finds oneself easily restricted by the default approach. This information however is essential in multi-user scenarios in order to properly interpret the user input. Since the traditional approach does not provide any further information other than the contact point itself additional processing is required to establish that information which obviously increases the computational cost of the pipeline.

Besides the restrictions on interaction metaphors further constraints are applied on the allowed hand posture during user interaction. The traditional approach usually requires the fingertips to be the nearest body parts to the surface in order to ensure proper functioning and avoid unintentional touch points. Obviously expert users familiar with the functioning automatically adopt that posture while novice users might encounter those pitfalls discouraging them in their further use. However keeping fingers in an upright angle during multi touch interaction leads to strain and fatigue in long term usage. In this case the user would normally want to rest the palm or the whole hand on the surface however hereby disturbing the touch detection and possibly triggering unintended behavior.

Similarly to resting the hand on the surface any other object placed onto the table might cause the same unwanted behavior. This however fundamentally contradicts the metaphor of a table which would allow placing objects such as tools necessary during the work process onto the table. Ideally the processing pipeline would differentiate between objects that are in direct interaction with the interactive tabletop such as those with an optical marker or unrelated such as pen and paper.

\begin{figure}[t]
	\centering
    \includegraphics[width=\textwidth]{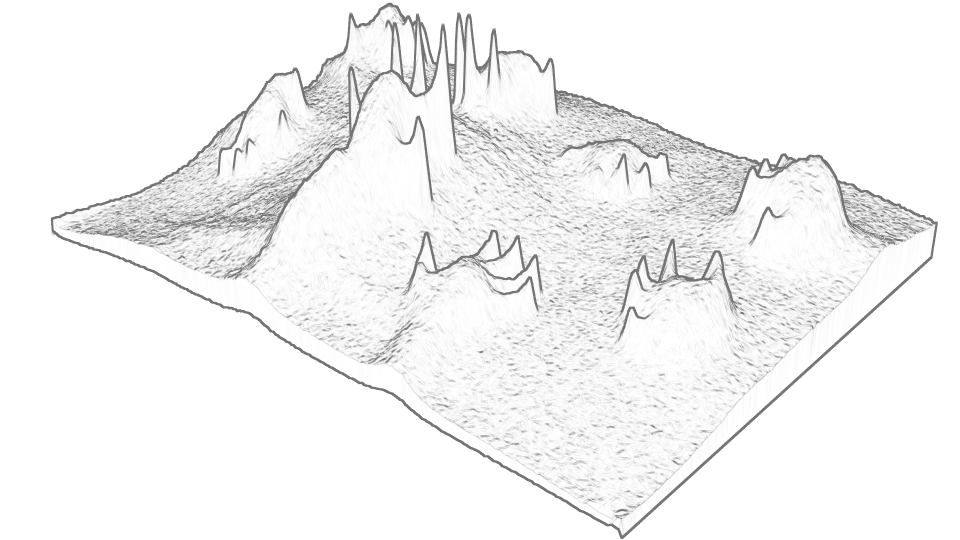}
    \caption{Camera image visualized as intensity height map. The uneven illumination is reflected in the deformed ground plane as well as in the highly unequal light response of all the fingers currently touching the surface.}
    \label{fig:prototype_illumination}
\end{figure}

Additionally to the aforementioned shortcomings when it comes to user interaction, from a technical viewpoint the assumption of uniform illumination which is essential for the traditional processing pipeline can be said to be impossible to achieve. Especially if tabletop displays grow in size or if only limited space is available behind the surface illumination properties tend to vary significantly across the surface as described in \cite{gokcezade2010lighttracker}. The same holds true for the presented prototype as can be seen in figure \ref{fig:prototype_illumination}.

This thesis proposes a new approach to contact point detection that not only identifies objects on the surface but additionally provides spatial and hierarchical information while being robust in the presence of non-uniform illumination. Instead of processing the camera image with image filters and thresholding to detect objects in contact with the surface, this approach is based on the analysis of distinguished regions in the camera image. At the core of the processing pipeline is an algorithm introduced by \citeauthor{matas2004robust} called Maximally Stable Extremal Regions that reveals the hierarchical structure of extremal regions in an image. An extremal region\footnote{see page \pageref{extremal_region} for a mathematical definition. In their original paper the concept of extremal region also includes the inverted case that all pixels inside a region are of lower intensity than those on the region's boundary. However this case is not relevant here and has therefore been omitted for the sake of simplicity.} is a particular type of distinguished region and can be defined as a set of connected pixels that are all of higher intensity than pixels on the region's outer boundary. Given the fact that objects in diffuse front illumination setups appear brighter the closer they get to the surface the representation of image content in terms of extremal regions provides some compelling advantages:
\begin{itemize}
\item An extremal region only defines a relative relationship between the contained pixels and those surrounding the region. Since it is independent from absolute intensity values it is more robust in the presence of non-uniform illumination than approaches relying on a global threshold.
\item One or more extremal regions can again be included in a larger extremal region hereby organizing distinguished image regions in a tree structure. That structure is later used to reveal relationships between objects such as grouping contact points belonging to the same hand.
\item Extremal regions can be considered a very reliable and stable object representation as they have been successfully used as object features in visual tracking in \cite{donoser2006efficient}.
\end{itemize}

\begin{figure}[p]
	\centering
	\includegraphics[width=0.8\textwidth]{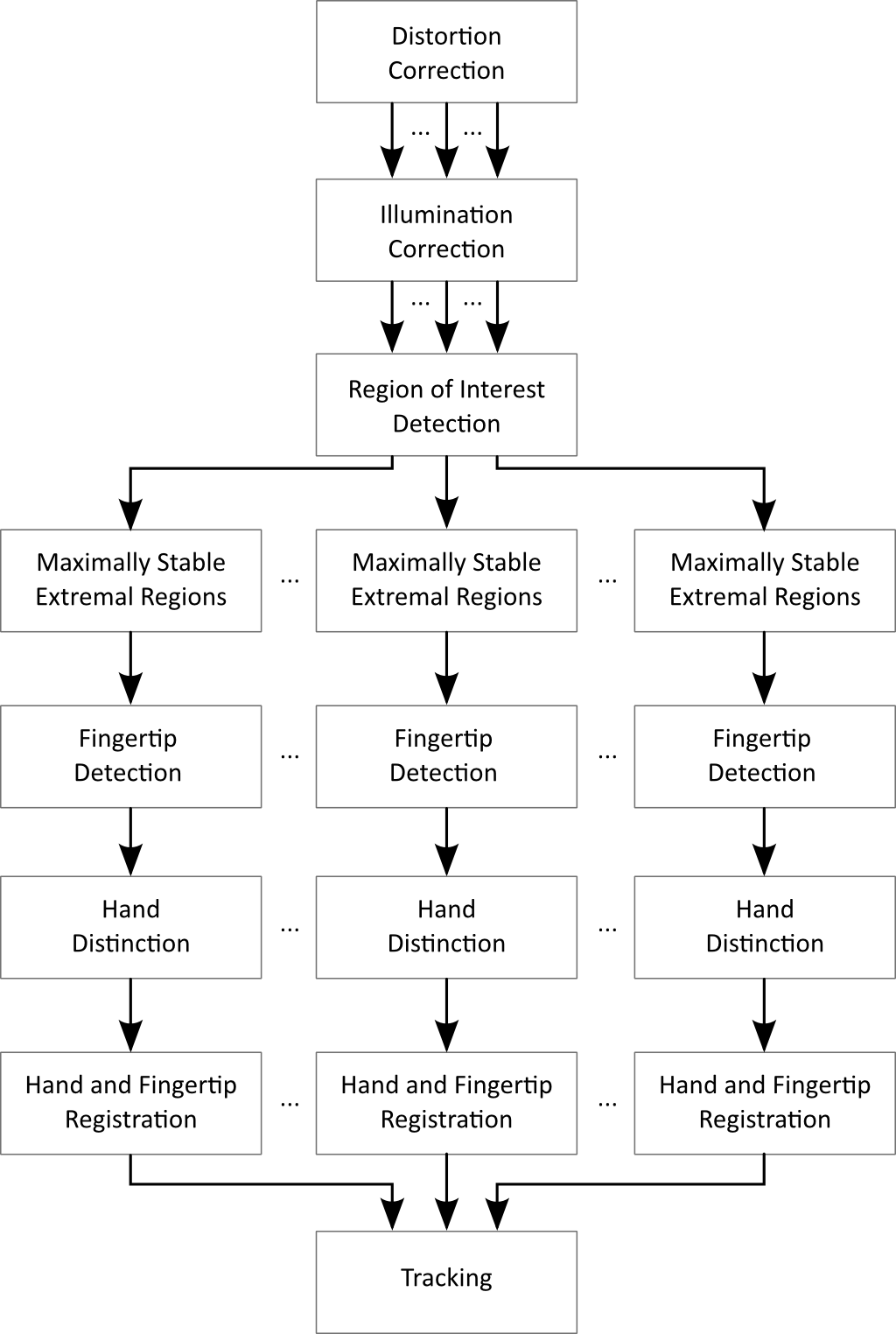}
    \caption{Diagram of the processing pipeline. The arrows illustrate the use of concurrency within the pipeline.}
    \label{fig:pipeline}
\end{figure}

The complete processing pipeline is illustrated in figure \ref{fig:pipeline} and comprises the following steps.
\begin{description}
\item[Distortion Correction] Reduce the distortion inherent to the camera optics and transform the resulting image such that the multi-touch surface covers the whole image.
\item[Illumination Correction] Reduce the effects of ambient illumination by normalizing intensity values across the image.
\item[Region of Interest Detection] Detect candidate regions for further processing to reduce the computational cost of subsequent processing steps.
\item[Maximally Stable Extremal Regions] Analyze the regions of interest for extremal regions and create the hierarchical structure.
\item[Fingertip Detection] Analyze the hierarchical structure and a number of image features computed from the extremal regions to identify fingertips touching the surface.
\item[Hand Distinction] Group all revealed fingertips from the same hand into clusters.
\item[Hand and Fingertip Registration] If all fingers of a hand are simultaneously touching the surface, a hand and fingertip registration process is performed. The process distinguishes between left and right hands and maps the revealed fingertips to the five fingers of a hand.
\item[Tracking] Establish intra-frame correspondences between hands and fingertips from consecutive frames.
\end{description}

\subsection{Preprocessing}
The preprocessing steps transform the incoming camera image into a normalized form in order to reduce image variations that would negatively impact performance of the following processing steps. These steps take into account both spatial and intensity transformations of the camera image that stem from intrinsic parameters of the prototype such as image distortions due to the camera and extrinsic parameters such as ambient illumination.

\subsubsection{Distortion Correction}
In order to evaluate input on the multi-touch surface, a uniform and undistorted view of the surface is required. However the representation of the surface in the image acquired from the camera has been subject to a number of deteriorating effects:
\begin{description}
\item[Barrel Distortion] \hfill\\
Since the employed camera uses a fisheye lens to capture the multi-touch surface from short range, it inevitably introduces \textit{barrel distortion}. This type of distortion is characterized by a bulging of usually straight lines resulting in a magnification effect of objects towards the image center.
\item[Perspective Transformation]\hfill\\
As the camera views the surface from a non-orthogonal angle, the image representation of the surface has undergone perspective transformation. Hence the surface appears to decrease in size with increasing distance.
\end{description}

Moreover it is in the end solely the image part covering the surface that is of any interest to us. Hence an image transformation is to be found that rectifies that image part while correcting the aforementioned effects as well.

\begin{figure}[tp]
	\centering
    \subfloat[]{\label{fig:undistortion_before}\includegraphics[width=0.4\textwidth]{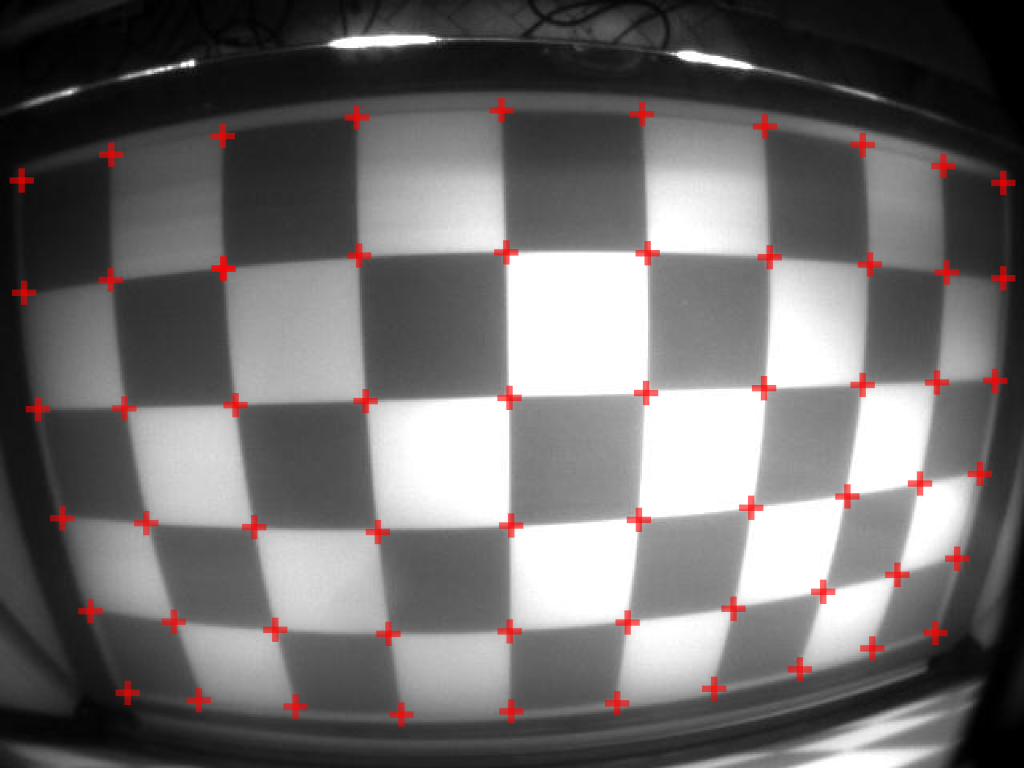}}
    \hspace{1cm}
    \subfloat[]{\label{fig:undistortion_after}
	    \raisebox{0.5cm}{\includegraphics[width=0.4\textwidth]{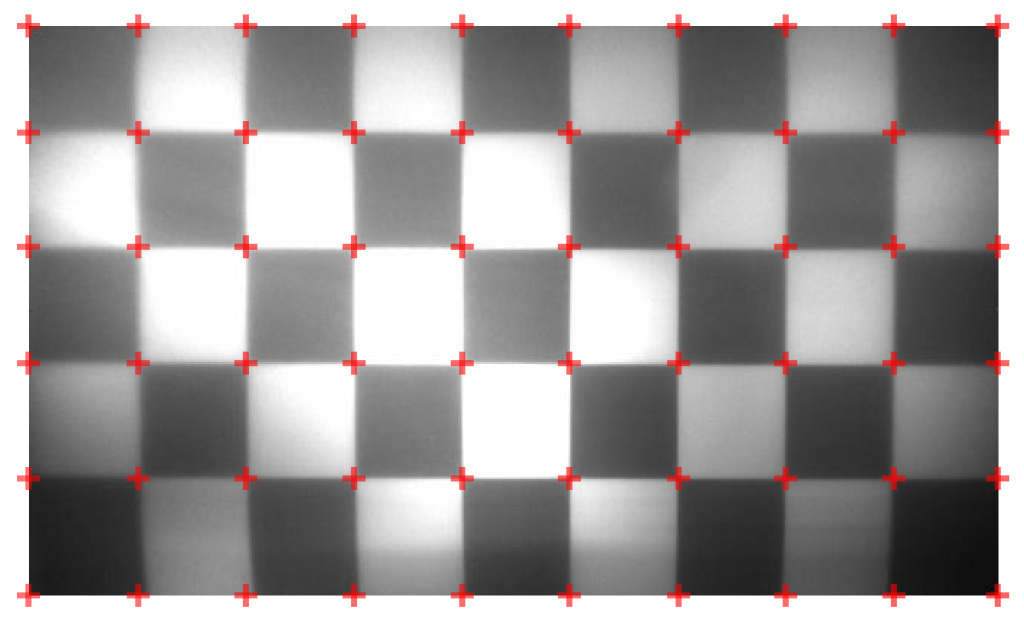}}}
    \caption{Illustration of the set of grid points in the distorted and undistorted image.}
    \label{fig:undistortion}
\end{figure}

The usual approach is to define a set of points in the image that cover the entire surface and are known to have formed a rectangular grid prior to being distorted. This approach has the advantage of ignoring the origin of the distortions and reliably works in our scenario. The grid points have been defined in our prototype by covering the surface with a printed checkerboard pattern where the corners of black and white squares represent the rectangular grid (see figure \ref{fig:undistortion}). Contrarily to what one might think, the next step is not to find a transformation that maps points from the distorted to the undistorted set, but to compute the inverse transformation that allows us to find for each pixel in the undistorted image its exact correspondence in the distorted image. 

Be $P_{m,n}$ and $Q_{m,n}$ the positions of the points in the distorted and undistorted images respectively. Since we know that the distorted points used to form a rectangular grid, a two-dimensional uniform bicubic B-spline interpolation will be used that defines a mapping between the parametric coordinates $u,v$ and their distorted position $(x,y)$ as follows:
\begin{equation}
f_2(u,v) = (x,y) = \sum\limits^{2}_{i=-1}\sum\limits^{2}_{j=-1}B_{i+1}(u)B_{j+1}(v)P_{m+i,n+j}
\end{equation}
where $(u,v)$ are the parametric coordinates inside the grid cell formed by the four points $Q_{m,n},Q_{m+1,n},Q_{m+1,n+1},Q_{m,n+1}$. Cubic B-spline interpolation defines four additional blending functions $B_i$ that determine the weight for each point and are defined as follows:
\begin{equation}
\begin{array}{ccc}
B_0(u)&=& \frac{1}{6} (1-u)^3\\
B_1(u)&=& \frac{1}{6} (3u^3-6u^2+4)^3\\
B_2(u)&=& \frac{1}{6} (-3u^3+3u^2+3u+1)\\
B_3(u)&=& \frac{1}{6} u^3\\
\end{array}
\end{equation}

Since the resulting distorted coordinates most likely fall in between actual pixels in the camera image, bilinear filtering on the intensity value is employed to interpolate between neighboring pixels. Obviously in a real-time processing pipeline the above bicubic B-spline interpolation would not be computed again at every frame. Unless the internal setup changes, the mapping between distorted and undistorted remains constant and only needs to be computed once.

\subsubsection{Illumination Correction}
As detailed in section \ref{background_subtraction} illumination correction is an important step in the processing pipeline in order to achieve accurate results. While usually only \textit{background subtraction} is performed more care needs to be taken here due to the following specificities of this setup that influence the illumination of a point on the tabletop surface:

\begin{description}
\item[Tilting Effects]
The tabletop surface is tilted with respect to the camera in this setup. Usually the camera is aligned such that camera axis and surface plane are at a right angle. This reduces perspective distortion effects however in order to preserve the compactness of the previously developed prototype this has not been changed. While perspective distortion can be easily corrected a more important effect is the intensity fall off resulting from the tilted surface. 

Light rays reflected from objects on and above the surface are transmitted to the camera passing through the acrylic surface. At both sides of the surface these rays are subject to reflection and refraction. According to Snell's Law the incident angle and the refracted angle of a light ray at an interface are proportionally related. Hence the more oblique the angle between camera and surface the more deviates the incident angle from the surface normal. However as the incident angle determines the amount of internal reflection\footnote{The amount of internal reflection rises with increasing incident angle until a critical angle that depends on the adjacent materials of an interface. Beyond the critical angle all light is reflected inside a material resulting in \textit{total internal reflection}. This phenomenon is exploited in FTIR multi touch setups as described on page \pageref{ftir}.} inside the surface panel one can conclude that camera angle and intensity fall off are directly related. 

\item[Vignetting Effects]
Wide-angle cameras such as the one used in this setup tend to be affected by vignetting effects. These describe non-uniform transformations in the image formation process due to several mechanisms related to the camera system. An often observed phenomenon is the irradiance fall off in the image periphery which can be attributed to vignetting. Vignetting occurs when the beam of incident light is blocked by internal parts of the optical system which is especially relevant for light arriving at oblique angles at the lens. Vignetting is prominent for larger apertures and generally absent for smaller apertures \cite{aggarwal2002cosine}. Another aberration however less important than vignetting is cosine-fourth. Cosine-fourth law describes an off-axis intensity fall off due to the foreshortening of the lens with respect to a point in space \cite{zheng2009single}. Another phenomenon is pupil aberration resulting in a non-uniform intensity distribution due to non-linear refraction of incoming rays \cite{aggarwal2002cosine}.
\end{description}

Considering these effects it is obvious that a simple shifting of intensity values on a per-pixel basis as in \textit{background subtraction} is not sufficient. Assuming that intensity variations due to the aforementioned effects can be represented for each pixel as affine transformations the relationship between original ($I'(x,y)$) and captured intensity values ($I(x,y)$) can be described as follows:
\begin{equation}
I(x,y) = a + m \cdot I'(x,y)
\end{equation}
with $a$ and $m$ being the additive and multiplicative component respectively ($a$ and $m$ are also referred to as \textit{noise} and \textit{intensity inhomogeneity effect} respectively).

\begin{figure}[tp]
	\centering
    \subfloat[]{\label{fig:normalization_before}\includegraphics[width=\textwidth]{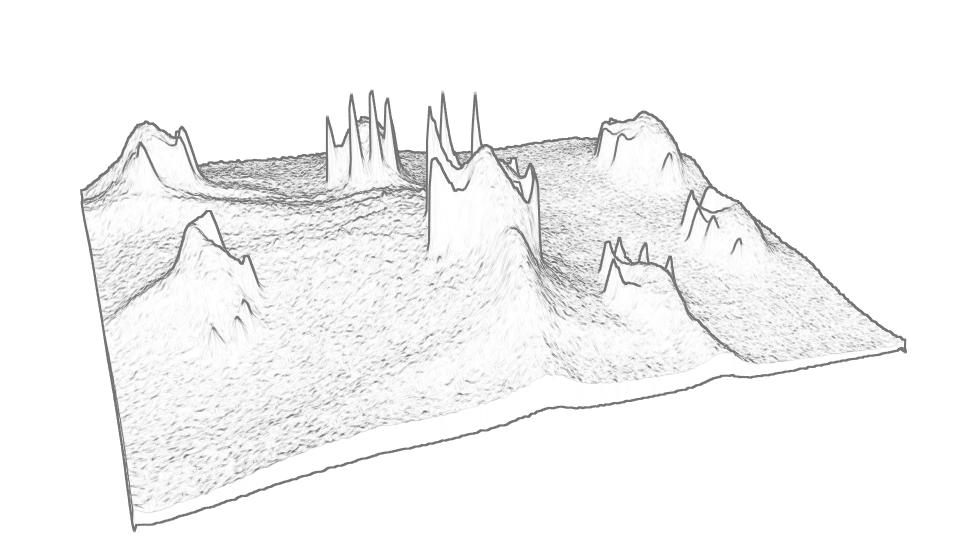}}
    \hspace{1cm}
    \subfloat[]{\label{fig:normalization_after}\includegraphics[width=\textwidth]{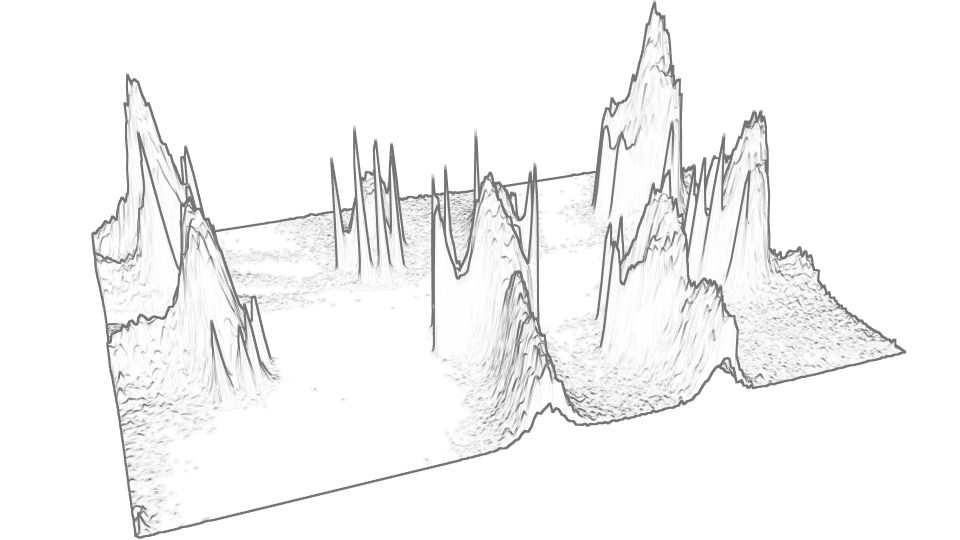}}
    \caption{Camera images before and after the normalization step visualized as intensity height maps. Fingertips are clearly visible as small spikes while palms and arms are larger in size and have a rather blunt appearance.}
    \label{fig:normalization}
\end{figure}

As a trade-off between performance and accuracy a min-max normalization approach was chosen using precomputed minimum and maximum intensity images. Similarly to \textit{background subtraction} the minimum intensity values are defined by a prerecorded background image. Reasonable maximum intensity values can be found by moving a sheet of paper across the surface and to recording the maximum intensity values for each pixel. Given a maximum intensity image $I_{max}$ and a minimum intensity image $I_{min}$ the normalized intensity value $I_{normalized}$ is computed by
\begin{equation}
I_{normalized}(x,y) = \frac{I(x,y) - I_{min}(x,y)}{I_{max}(x,y) - I_{min}(x,y)}\cdot 2^b
\end{equation}
with $b$ representing the image \textit{bit depth}.
The maximum intensity image depends solely on the internal configuration of the prototype, hence requiring no recalibration unless the setup changed. The minimum intensity image similarly to the static \textit{background subtraction}\footnote{see page \pageref{background_subtraction}} approach requires recalibration if the external lighting situation changes significantly. However the subsequent processing steps have been found to be robust enough to be unaffected by even significant illumination changes due to stray sun light. Illumination changes during operation might in some circumstances result in $I(x,y) < I_{min}(x,y)$ or $I(x,y) > I_{max}(x,y)$. Therefore the value $I_{normalised}(x,y)$ needs to be clamped to the range $(0,2^b($ to ensure proper functioning in the following stages. The effect of this normalization steps is visualized in figure \ref{fig:normalization}.

\subsubsection{Region of Interest Detection}\label{section_connected_components_horn}

Considering that users usually only interact with a limited area of the tabletop surface at a time it seems like an unnecessary overhead to process the complete camera image at every frame. Therefore a simple region of interest detection is performed that provides the following advantages:
\begin{itemize}
\item The following fingertip detection algorithm is limited to the relevant image areas that might represent an object or a hand on or above the surface hence significantly reducing computational costs.
\item This processing step results in a set of independent regions of interest. Given that independence those regions can be processed in parallel therefore taking full advantage of the threading capabilities of modern CPUs.
\end{itemize} 

As the tabletop surface is illuminated from below objects above the surface appear brighter the closer they get to the surface. The previous normalization step transformed pixel intensities such that a pixel's intensity is zero if it does not represent an object or larger than zero the closer it gets to the surface. Hence a simple \textit{thresholding}\footnote{see page \pageref{thresholding} for a detailed description of the \textit{thresholding} operation} operation on the intensity values seems appropriate to reveal regions of interest in the image. Although this approach is sensitive to illumination changes erroneously interpreting these intensity variations as user interaction, this only reduces the performance advantage without affecting the actual detection accuracy.

Thresholding of an image $I$ is defined as
\begin{equation}
T(x,y) = \begin{cases}
1\quad\mbox{if }I(x,y) \geq T  \\
0\quad\mbox{otherwise} 
\end{cases}
\end{equation}
with $T$ being the threshold. 

So far areas of interest in the camera image have been established on a per pixel basis only. Therefore pixels are now being grouped in distinct clusters according to a neighborhood criterion (throughout the processing pipeline a 4-neighborhood will be used). Those clusters are usually referred to as \textit{connected components}. Generally a connected component represents a region in a binary image for which all its pixels have an intensity of 1. However a broader definition of such a component will be used here:

\begin{description}
\item[Connected Component]\hfill\\
A region $R$ in an image $I$ represents a connected component $C$ if and only if
\begin{equation}
C = \{p\mid p\in R \wedge H(p, I(x_p, y_p)) = 1\}
\end{equation}
with $H$ being a homogeneity criterion function. $H$ returns $1$ if the criterion holds for a pixel or $0$ otherwise.

Thresholding represents the simplest form of a homogeneity criterion that is defined as
\begin{equation}
H(p, I(x_p, y_p)) = T(x_p, y_p)
\end{equation}
\end{description}

Connected component algorithms seek to find all connected components $C_1,\dots,C_n$ in an image $I$ such that these components are maximal, i.e. no more neighboring pixels can be added to this component:
\begin{equation}
\forall p\in \Omega_{C_i} \mid H(p, I(x_p, y_p)) = 0
\end{equation}
The simplest of all connected component algorithms works by starting at an arbitrary pixel and to recursively scan the image (see section \ref{section_connected_components}). However as the component grows randomly in all directions it hardly exploits any cache locality since images are generally stored sequentially. Therefore a sequential connected components algorithm will be used here as proposed by \citeauthor{horn1986robot} in \cite{horn1986robot}.

\begin{figure}[t]
	\centering
    \subfloat[]{\label{fig:horn_cc_mask}
    \raisebox{1cm}{\includegraphics[width=0.1\textwidth]{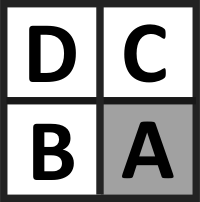}}}
    \hspace{2cm}
    \subfloat[]{\label{fig:horn_cc_merge}\includegraphics[width=0.4\textwidth]{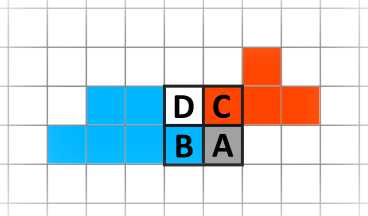}}
    \caption{(a) Pixel mask used in the connected components algorithm.\\(b) Case where two previously disconnected components will be merged.}
\end{figure}

The algorithm scans the image row-by-row, i.e. left-to-right and top-to-bottom, and therefore accesses image pixels in the same order as they are aligned in memory. When a pixel is accessed it is evaluated according to the defined homogeneity criterion and a label representing the connected component is added to the pixel. If the criterion does not hold true the label is simply zero. Otherwise neighboring pixels need to be considered in order to decide on a label for the current pixel. During execution all pixels above and left from the current pixel are known to have been evaluated before. As it is known to which components these pixel belong, the 2x2 matrix as shown in figure \ref{fig:horn_cc_mask} is sufficient to label a pixel. Depending on the previously labeled pixels, the following cases may arise:
\begin{enumerate}
\item \textbf{Neither B,C or D have been labeled}\hfill\\ \label{conn_comp_case_1}
Create a new label for A
\item \textbf{B or C has been labeled}\hfill\\
Label A according to the label of B or C
\item \textbf{B and C have been labeled}\hfill\\\label{conn_comp_case_3}
Two cases need to be considered:
\begin{itemize}
\item If B and C have the same label then just copy this label.
\item If B and C have not the same label then the equivalence of these two labels is recorded. The label of B or C may be copied to A. (see figure \ref{fig:horn_cc_merge} as illustration of this case)
\end{itemize}
\item \textbf{Only D has been labeled}\hfill\\
If using a 8-neighborhood simply copy the label from D otherwise go to case~\ref{conn_comp_case_1}
\end{enumerate}

After the first processing pass the whole image has been processed and each pixel has been assigned a label representing its connected component. However due to case \ref{conn_comp_case_3} there might exist a number of pixels that all belong to the same component however have been assigned different labels. Therefore in the original algorithm a second processing pass is being performed to achieve a consistent pixel labeling. During this pass all equivalent labels would be replaced with a new unique label.

The following processing steps are also based on a 4-neighborhood criterion and require the image labeling as an accessibility mask only. Since detected regions are by definition not connected, a consistent labeling is not required and the second processing pass can be omitted here. However a number of different properties is computed for each region during the first processing pass:
\begin{description}
\item[Pixel Count] The pixel count will be used to discard regions that are too small and are considered to originate from noise.
\item[Intensity Histogram] The intensity histogram will be used to efficiently organize data structures used during the execution of Maximally Stable Extremal Regions.
\item[Brightest Pixel] The brightest pixel of the region will be used as a starting point for the subsequent Maximally Stable Extremal Regions algorithm. Choosing that pixel as a starting point slightly increases the performance of the first steps of the algorithm.
\end{description}
In order to keep track of these features the label assigned to a pixel uniquely identifies a data structure containing this information. As regions, as in figure \ref{fig:horn_cc_merge}, might merge during the algorithm the data structure additionally carries a reference to its root structure into which the new information will be merged. The root data structure is by definition the one that started a region, hence has been created earliest (the red region in the case of figure \ref{fig:horn_cc_merge}). Consider that the blue and red regions from the figure have been merged. If subsequently another region is to be merged into the blue one, it is going to be merged into the root structure instead (belonging to the red region), as the reference has been set during the previous merging.

\subsection{Maximally Stable Extremal Regions}

Maximally Stable Extremal Regions (MSER) is a widely used blob detection algorithm first described by \citeauthor{matas2004robust} in \cite{matas2004robust}. It robustly detects distinguished regions, i.e. regions that can be detected in image sequences or multiple views of the same scene with high repeatability. They then used these regions in order to establish correspondences in stereo image pairs. In the proposed algorithm they introduced a new type of distinguished region called maximally stable extremal region. This concept is an extension of extremal regions which are defined as follows:

\begin{description}
\item[Extremal Region]\hfill\\ \label{extremal_region}
Given a region $R$ in an image $I$, this region is called extremal if and only if all pixels within this region are either of higher or lower intensity than the pixels on the region's outer boundary:
$$
\left(\forall p \in R,\, \forall q \in \Omega_R \mid I(x_p, y_p) < I(x_q, y_q)\right)\, \vee \,
\left(\forall p \in R,\, \forall q \in \Omega_R \mid I(x_p, y_p) > I(x_q, y_q)\right)
$$
\end{description}

The concept of maximally stable extremal region additionally includes a stability criterion based on a region's relative growth.

\begin{description}
\item[Maximally Stable Extremal Region]\hfill\\
Be $R_{i-1}, R_{i}, R_{i+1}$ regions in an image $I$ such that $R_{i-1}\subset R_{i}\subset R_{i+1}$. Region $R_i$ is a maximally stable region if and only if $R_{i-1}, R_{i}, R_{i+1}$ are extremal regions and the following stability property attains a local minimum in $i$:
$$
s(i)=\frac{\mid R_{i+1}\mid - \mid R_{i-1}\mid}{\mid R_i \mid}
$$
In order to adjust the required level of contrast the criterion is usually extended to contain a user-defined constant $\Delta$:
$$
s_{\Delta}(i)=\frac{\mid R_{i+\Delta}\mid - \mid R_{i-\Delta}\mid}{\mid R_i \mid}
$$
\end{description}

\citeauthor{matas2004robust} describe the detection process of these regions informally as follows. Be $I$ a gray-level image thresholded on all possible intensity values $i\in\{0,\dots,255\}$ resulting in 256 binary images $B_i$. $B_0$ contains only black pixels while in $B_{255}$ all pixels are white. Hence, when iterating through $B_i$ with ascending $i$ the initial black image gradually turns into a complete white image with new white blobs appearing and existing blobs increasing in size. Considering the connected components from all thresholded images, each of them corresponds to an extremal region. Furthermore each extremal region $R_i$ at intensity threshold $i$ is contained by exactly one extremal region from each $B_j$ with $j > i$, since connected components do not decrease in size with increasing intensity threshold.

\begin{figure}[tp]
	\centering
    \subfloat[]{\includegraphics[width=0.7\textwidth]{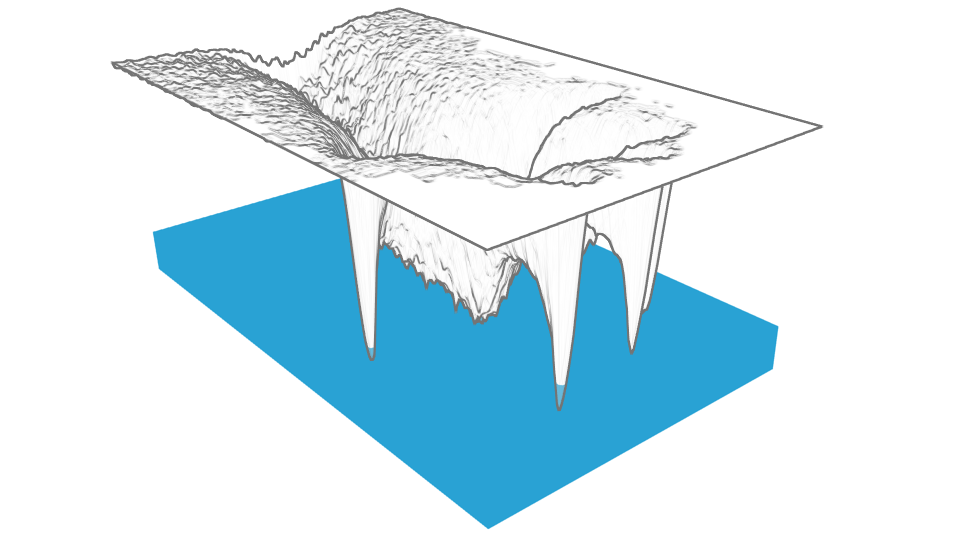}
			    \includegraphics[height=0.4\textwidth]{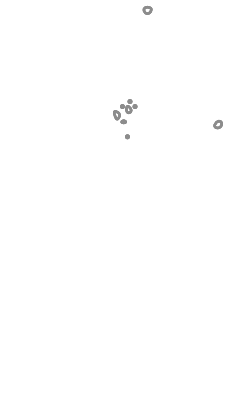}}
    \hspace{1cm}
    \subfloat[]{\includegraphics[width=0.7\textwidth]{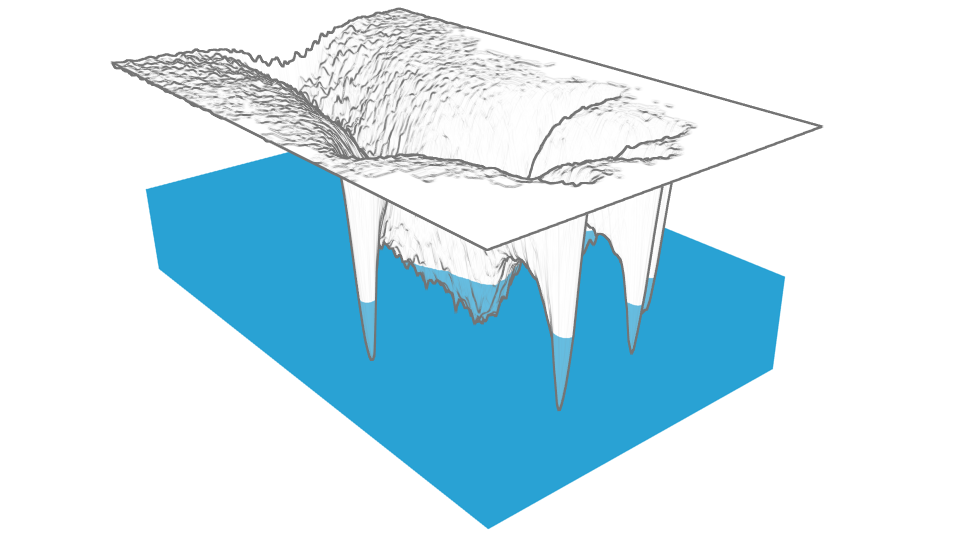}
			    \includegraphics[height=0.4\textwidth]{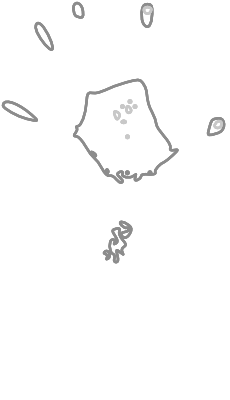}}
    \hspace{1cm}
    \subfloat[]{\includegraphics[width=0.7\textwidth]{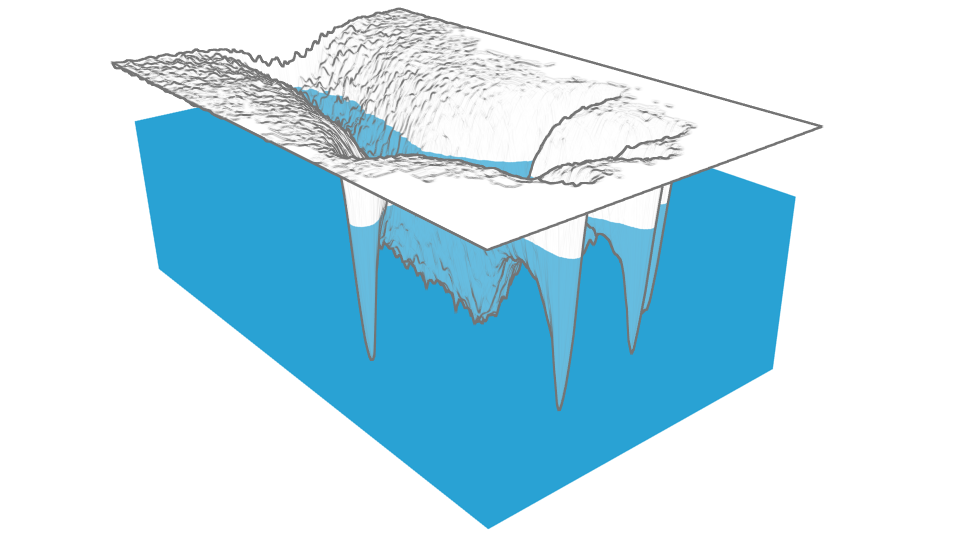}
			    \includegraphics[height=0.4\textwidth]{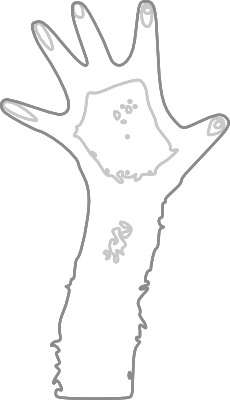}}
    
    \caption{Visualization of the \textit{Maximally Stable Extremal Region} algorithm as a rising waterfront. Left: The image represented as intensity height map where the waterfront indicates the current threshold. Right: The outline of the thresholded image at the water level. Outlines from previous thresholds have been overlaid.}
    \label{fig:mser_threshold}
\end{figure}

The algorithm can be illustrated as shown in figure \ref{fig:mser_threshold} and exhibits a certain resemblance to a watershed segmentation\footnote{See \cite{roerdink2000watershed} for a complete description of the watershed transform}. Although the algorithmic design is similar the way regions are selected differs significantly. The set of regions returned by the watershed algorithm forms a partition of the whole image whereas regions revealed by MSER might overlap and usually only cover a limited areas of the image. However they mainly differ in when a region is considered interesting. Watershed segmentation focuses on threshold levels when previously distinct regions merge. These are highly unstable and therefore inapt as distinguished regions. In contrast MSER chooses regions at threshold levels where the size of the region remains unchanged within a defined intensity range.

\begin{figure}[t]
	\centering
	\includegraphics[width=\textwidth]{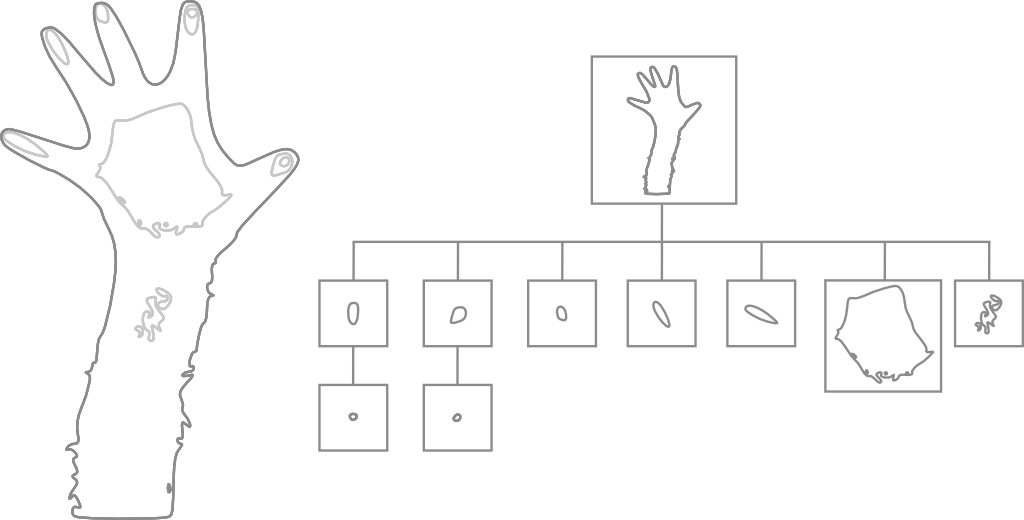}
    \caption{Illustration of a simplified component tree. Children of the root node represent from left to right the five fingers, the palm and the wrist.}
    \label{fig:mser_component_tree}
\end{figure}

As described above an extremal region at threshold level $i$ is contained by exactly one extremal region at each higher intensity level. However not all extremal regions are maximally stable, therefore it follows that a maximally stable extremal region at threshold level $i$ is included in at most one maximally stable extremal region from each $B_j$ with $j > i$. Hence, maximally stable extremal regions can be linked according to a parent - child relationship. Furthermore as regions merge with increasing threshold intensity a parent can have more than one child, however a child region has at most one direct parent region. Thus maximally stable extremal regions can be associated in a hierarchical structure called \textit{component tree}. Such a component tree is displayed in figure \ref{fig:mser_component_tree}\\

According to \citeauthor{matas2004robust} maximally stable extremal regions are characterized by the following properties:
\begin{enumerate}
\item Illumination invariant as they are invariant to affine transformation of image intensities
\item Viewpoint independent as they are covariant to adjacency preserving transformations on the image domain
\item Stability due to the stability criterion
\item Multi-scale detection
\end{enumerate}

\begin{description}
\item[Algorithm]\hfill\\
The algorithm as described in \cite{matas2004robust} proceeds as follows:
\begin{enumerate}
\item 
The algorithm can identify either extremal regions of lower intensity than its boundary pixels or those that are of higher intensity, however not both at the same time. Hence:
\begin{enumerate}
\item In the former case continue the execution of the algorithm normally.
\item In the latter case invert the source image hereby converting between the two types of extremal regions. This is the relevant case in this processing pipeline.
\end{enumerate}
\item
Pixels are sorted by ascending intensity. As intensity values are bound to the range from 0 to 255, they propose \texttt{BinSort} for this step which has a computational complexity of $\mathcal{O}(n)$ with $n$ being the number of pixels.
\item
Iterate through all intensity values, i.e. from 0 to 255 and add the pixels corresponding to the current intensity value to the image.
\item
Keep track of newly appearing, growing and merging connected components within the image. This is done using a union-find data structure\cite{tarjan1984worst} with quasi-linear computational complexity\footnote{The complexity is reached when using union-find with compression and linking by size. $\alpha(n)$ represents the inverse of the Ackermann function which is small for all practical $n$} of $\mathcal{O}(n\alpha(n))$ with $n$ being the number of pixels.
\item
Growth rate of connected components is evaluated according to the stability criterion and if locally minimal the connected component is selected as maximally stable extremal region.
\end{enumerate}
\end{description}

\begin{figure}[t]
	\centering
    \subfloat[]{\includegraphics[width=0.4\textwidth]{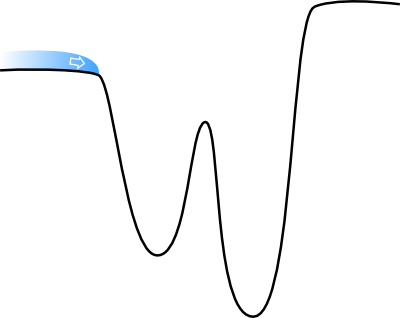}}
    \subfloat[]{\includegraphics[width=0.4\textwidth]{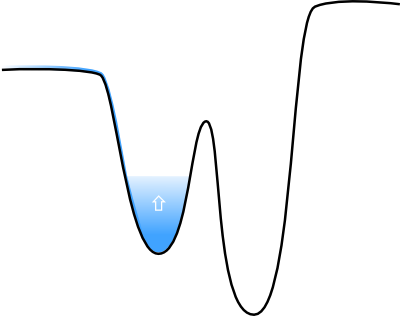}}\\
    \subfloat[]{\includegraphics[width=0.4\textwidth]{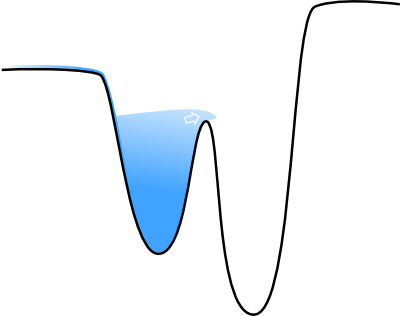}}
    \subfloat[]{\includegraphics[width=0.4\textwidth]{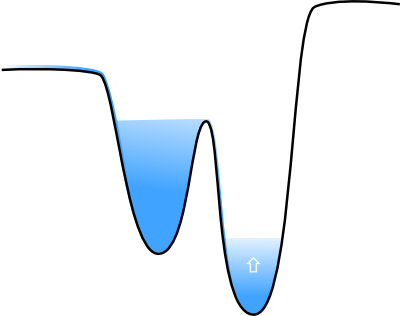}}
    \caption{Illustration of the linear time algorithm. (a) Water is poured at an arbitrary point and runs downhill until reaching a sink. (b) The sink is being filled with water. (c) Water flows over to a neighboring sink. (d) The neighboring sink is being filled with water.}
    \label{fig:mser_linear}
\end{figure}

Hence the algorithm runs in quasi-linear time with respect to the number of pixels. However in \citeyear{nistér2008linear} \citeauthor{nistér2008linear} proposed a different algorithm that finds maximally stable extremal regions in true linear time \cite{nistér2008linear}. Unlike the above algorithm it does not resemble a rising waterfront continuously filling holes in the intensity height image but rather a flood-fill spreading across the image (see figure \ref{fig:mser_linear}). Since only pixels on the boundary of the ''flooding'' are considered the algorithm does not rely on the union-find data structure as a simple priority queue is sufficient.

Informally speaking the algorithm can be outlined as follows: Considering an image as a height map based on its intensities. Water is poured on an arbitrarily chosen point from which water starts flowing downhill. Once the lowest point of a sink is reached water starts filling up. The filling of a sink would correspond to the growing of a connected component. Whenever the water surface of the sink did not change notably in size after filling a certain amount of water this connected component can be considered stable according to the above definition. Once a sink is fully filled water starts pouring into adjacent sinks. The algorithm is finished after the whole image has been flooded.

\begin{description}
\item[Linear-time algorithm]\hfill\\
The algorithm relies on the following data structures that describe the current state of execution:
\begin{itemize}
\item
A binary mask in order to mark the pixels that have already been visited
\item
A priority queue of pixels on the outer boundary of the currently flooded image region. The queue is prioritized based on the pixels intensity values with lower intensities coming first.
\item
A stack of components. A component corresponds to a sink that is currently being filled with water. Therefore a component is not always necessarily an extremal region. The topmost component on the stack corresponds to the currently considered component. Whenever a sink is filled and adjacent sinks exist, a new component is pushed onto the stack and the neighboring sink is explored. Therefore the maximum stack size is equal to the number of gray levels in the image, i.e. at most 256.
\end{itemize}

The algorithm proceeds as follows \cite{nistér2008linear}.

\begin{enumerate}
\item
Choose an arbitrary starting point, make it the current pixel and mark it as visited.
\item \label{mser_algo_init_comp}
Initialize a new component with the current pixel's intensity and push it onto the stack.
\item\label{mser_algo_exam_neighbor}
Examine all neighboring pixel of the current pixel and mark them as visited. 
\begin{itemize}
\item
If the intensity is higher than the current pixel's intensity the neighboring pixel is added to the priority queue of boundary pixels. 
\item
If the intensity is lower the current pixel is added to the priority queue and the neighboring pixel is made the current pixel. Continue with \ref{mser_algo_init_comp}.
\end{itemize}
\item
All neighboring pixel of the current pixel have either been visited or have higher intensity. Therefore accumulate the component with the current pixel.
\item
Get the next pixel from the priority queue.
\begin{itemize}
\item
If its intensity is equal to the current pixel's intensity, continue with \ref{mser_algo_exam_neighbor}.
\item
If its intensity is higher than the current pixel's intensity, all components on the stack are merged until the intensity of the topmost component is greater or equal to the next pixel's intensity. Continue with \ref{mser_algo_exam_neighbor}.
\item
If the queue is empty, the algorithm terminates.
\end{itemize}
\end{enumerate}
\end{description}

\paragraph{Modifications of the algorithm}
In order to accelerate the subsequent processing steps, the algorithm has been modified to gather additional information on extremal regions and their spatial structure during its execution. 
\begin{description}
\item[Extension to reveal all extremal regions]\hfill\\
In the default design the algorithm only reveals extremal regions that are maximal according to the stability criterion. Hence a number of intermediate extremal regions are being discarded and do not appear in the component tree. However these additional regions might convey important information as to the spatial relationship between regions that will be exploited extensively in the hand distinction processing step. Since this information is considered to be crucial, the algorithm design has been changed to identify all available extremal regions. Nonetheless the stability measure is still being computed for each region and might be used in further processing steps.
\item[Characterization of extremal regions using local descriptors]\hfill\\
In order to being able to use the identified regions for fingertip recognition a set of local descriptors will be used. Due to the incremental growth of components the following features can be efficiently computed during execution of the algorithm. Be $R$ the region defined by the component in an image $I$:
\begin{description}
\item[Intensity statistics]\hfill\\
The mean and variance of the intensity distribution characterize the pixel's intensity values within a component. Generally, the mean $\mu$ and variance $\sigma^2$ (the variance is represented as the squared standard deviation here) for a discrete signal are defined as follows:
\begin{eqnarray}
\mu &=& \frac{1}{\mid R \mid} \sum_{p\in R} I(x_p, y_p) \\
\sigma^2 &=& \frac{1}{\mid R \mid} \sum_{p\in R} \left(I(x_p, y_p) - \mu\right)^2 \\
	 &=& \frac{1}{\mid R \mid} \sum_{p\in R} I(x_p, y_p)^2 - \left(\frac{1}{\mid R \mid} \sum_{p\in R} I(x_p, y_p)\right)^2\\
	 &=& \frac{1}{\mid R \mid} \sum_{p\in R} I(x_p, y_p)^2 - \mu^2
\end{eqnarray}
The mean is simply represented by the arithmetic mean of all pixel intensities within a region while the variance can be described as the mean of squares minus the square of the mean.

During the gradual growth of components through accumulation of new pixels and merging of existing components, these statistics can be be efficiently updated by keeping track of the following two equations:
\begin{eqnarray}
S_1 &=& \sum_{p\in R} I(x_p, y_p)\label{intensity_sum1} \\
S_2 &=& \sum_{p\in R} I(x_p, y_p)^2\label{intensity_sum2}
\end{eqnarray}

Whenever a pixel is added to the region the intensity and its square are added to the two equations. On merging the sums $S_1^{(1)}$ and $S_1^{(2)}$ and $S_2^{(1)}$ and $S_2^{(2)}$ from two regions $R^{(1)}$ and $R^{(2)}$ respectively are simply added one to another.

Based on equations \ref{intensity_sum1} and \ref{intensity_sum2} the mean $\mu$ and variance $\sigma^2$ are calculated as follows:
\begin{eqnarray}
\mu &=& \frac{S_1}{\mid R \mid} \\
\sigma^2 &=& \frac{S_2}{\mid R \mid} - \mu^2
\end{eqnarray}

\item[Image moments]\hfill\\
Image moments have been widely used in image processing algorithms for pattern recognition or object classification. A set of image moments can accurately describe an objects shape as well as a range of different geometric properties \cite{mukundan1998moment}. Furthermore image moments can be used to form a set of shape descriptors that are invariant with respect to a set of image transformations. \citeauthor{hu1962visual} was the first to use image moments in pattern recognition. In \cite{hu1962visual} he derived a still widely used set of descriptors that are invariant to translation, rotation and scale. 

Generally one distinguishes two types of moments in image processing, \textit{geometric} and \textit{photometric moments}. \textit{Geometric moments} only take into account the shape information of a region while \textit{photometric moments} include each pixel's intensity value as well.

The basic definition of image moments is as follows \cite{flusser2006moment}:\\
Given a real function $f(x,y)$ with finite non-zero integral, the two-dimensional moment of order $(p+q)$ with $p,q \in \mathbb{N}$ is defined as
\begin{equation}
m_{pq} = \int\int x^p y^q f(x,y)\mbox{dxdy}
\end{equation}

Considering the discrete case and a region $R$ for which the image moment is to be calculated, the above equation then becomes
\begin{equation}
m_{pq} = \sum_{(x,y)} x^p y^q f(x,y)
\end{equation}

The difference between \textit{geometric} and \textit{photometric moments} is in the definition of the function $f$. In the case of \textit{geometric moments} $f(x,y)$ would be
$$
f(x,y) = \left\{\begin{array}{cl}
1 & \mbox{ if $(x,y) \in R$}\\
0 & \mbox{ otherwise}
\end{array}\right.
$$
while in the case of \textit{photometric moments} $f$ would be defined as
$$
f(x,y) = \left\{\begin{array}{cl}
I(x,y) & \mbox{ if $(x,y) \in R$}\\
0 & \mbox{ otherwise}
\end{array}\right.
$$

The following part will focus on \textit{geometric moments} therefore $f$ is considered to be defined accordingly.

From $m_{pq}$ a certain number of characteristics may be derived. $m_{00}$ represents the number of pixels within the considered region, i.e. its area, while $x_c = m_{10}/m_{00}$ and $y_c = m_{01}/m_{00}$ define its center of mass $(x_c, y_c)$. The center of mass is the point where the whole mass of a region might be concentrated without changing its first moments \cite{horn1986robot}.
As $m_{pq}$ is invariant neither to translation, rotation nor scale, it will be called a \textit{raw moment}. In contrast \textit{central moments} are invariant to translation and defined as
\begin{equation}
\mu_{pq} = \sum_{(x,y)} (x-x_c)^p (y-y_c)^q f(x,y)
\end{equation}

\textit{Central moments} have the following characteristics:
$$
\mu_{00} = m_{00},\quad \mu_{10}=\mu_{01}=0,
$$
$\mu_{20}$ and $\mu_{02}$ represent the variance about the center of mass, while the covariance measure is defined by $\mu_{11}$ \cite{mukundan1998moment}.

Besides using the previously described equation, \textit{central moments} can also be calculated directly from \textit{raw moments}:
\begin{equation}
\mu_{pq} = \sum_k^p \sum_j^q \binom{p}{k} \binom{q}{j} (-x_c)^{(p-k)} (-y_c)^{(q-j)} m_{kj}
\end{equation}

\textit{Central moments} can be made scale invariant through normalization, therefore \textit{normalized moments} can be derived as follows:
\begin{equation}
\nu_{pq} = \frac{\mu_{pq}}{\mu_{00}^{(p+q+2)/2}}
\end{equation}

Having achieved translation and scale invariance so far, it was \citeauthor{hu1962visual} who first published shape descriptors that were also rotation invariant. The 7 shape descriptors $\phi_i$ as proposed by \citeauthor{hu1962visual} are as follows:
\begin{equation*}
\begin{array}{cccl}
\phi_1 &=& &\nu_{20} + \nu_{02} \\
\phi_2 &=& &(\nu_{20} + \nu_{02})^2 + 4 \nu_{11}^2 \\
\phi_3 &=& &(\nu_{30} - 3\nu_{12})^2 + (3\nu_{21} - \nu_{03})^2 \\
\phi_4 &=& &(\nu_{30} + \nu_{12})^2 + (\nu_{21} + \nu_{03})^2 \\
\phi_5 &=& &(\nu_{30} - 3\nu_{12})(\nu_{30} + \nu_{12})\left[(\nu_{30} + \nu_{12})^2 - 3(\nu_{21} + \nu_{03})^2\right] \\
       & &+&(3\nu_{21} - \nu_{03})(\nu_{21} + \nu_{03})\left[3(\nu_{30} + \nu_{12})^2 - (\nu_{21} + \nu_{03})^2\right] \\
\phi_6 &=& &(\nu_{20} + \nu_{02})\left[(\nu_{30} + \nu_{12})^2 - 3(\nu_{21} + \nu_{03})^2\right]\\
       & &+&4\nu_{11}(\nu_{30} + \nu_{12})(\nu_{21} + \nu_{03})^2 \\
\phi_7 &=& &(3\nu_{21} - \nu_{03})(\nu_{30} + \nu_{12})\left[(\nu_{30} + \nu_{12})^2 - 3(\nu_{21} + \nu_{03})^2\right]\\
       & &-&(\nu_{30} - 3\nu_{12})(\nu_{21} + \nu_{03})\left[3(\nu_{30} + \nu_{12})^2 - (\nu_{21} + \nu_{03})^2\right]
\end{array}
\end{equation*}

\item[Bounding volumes]\hfill\\
Image moment based shape descriptors provide a performant but unintuitive way of describing a region's shape, however it is sometimes useful to find a simpler representation based on geometric primitives. The simplest form of such a representation is an axis-aligned rectangle enclosing all pixels belonging to a region. Though being a very rough approximation it is well suited to locate a region in an image with respect to other regions. Such a rectangle with minimal area is called \textit{bounding box}.

A severe drawback of an axis-aligned \textit{bounding box} is its insensitivity to object orientation. The area of a bounding box might increase through object rotation, although the actual area of the object did not change. Therefore another shape representation using an oriented ellipse is presented. Different approaches to compute such an ellipse exist, however optimal techniques are computationally expensive. Nonetheless a non-optimal approximation can be easily derived using \textit{image moments}. Given a region $R$ with geometric moments $m_{00}, m_{10}, m_{01}, m_{11}, m_{20}, m_{02}$ its bounding ellipse can be computed as follows \cite{mukundan1998moment,rocha2002image}.

Since the resulting bounding ellipse conserves the source energy, i.e. has the same zeroth, first and second order moments, the centroid $(x_c, y_c)$ of the enclosing ellipse is equal to the center of mass of $R$:
\begin{equation}
x_c = \frac{m_{10}}{m_{00}} \quad\mbox{and}\quad y_c = \frac{m_{01}}{m_{00}}
\end{equation}
The orientation $\theta$, semi-minor axis $h$ and semi-major axis $w$ can be computed by
\begin{equation*}
\begin{array}{ccl}
\theta &=&\dfrac{1}{2}\cdot\tan^{-1}\left({\dfrac{b}{a-c}}\right) \\
h &=&\sqrt{2\cdot\left(a+c-\sqrt{b^2+(a-c)^2}\right)} \\
w &=&\sqrt{2\cdot\left(a+c+\sqrt{b^2+(a-c)^2}\right)}
\end{array}
\end{equation*}
with
\begin{equation*}
\begin{array}{ccccl}
a &=& \dfrac{\mu_{20}}{m_{00}} &=&\dfrac{m_{20}}{m_{00}} - x_c^2 \\
b &=& 2\cdot\dfrac{\mu_{11}}{m_{00}} &=&2\cdot\left(\dfrac{m_{11}}{m_{00}} - x_c y_c\right) \\
c &=&  \dfrac{\mu_{02}}{m_{00}} &=&\dfrac{m_{02}}{m_{00}} - y_c^2 
\end{array}
\end{equation*}
with $\mu_{20},\mu_{02}$ and $\mu_{11}$ being the second order central moments of $R$.
\end{description}

\end{description}

\subsection{Fingertip Detection}
The component tree resulting from the MSER processing is at the basis of this and the following processing steps. This tree structure has two important properties with respect to its contained \textit{extremal regions}:
\begin{itemize}
\item The darkest and at the same time largest extremal region forms the root of the component tree.
\item For all child-parent relationships holds the property that the child is smaller in size and has brighter intensity than its parent.
\end{itemize}
As fingertips in contact with the surface create the brightest spots in an image it follows that only leaf nodes of the component tree can be considered fingertip candidates.\\

These fingertip candidates are evaluated as follows:
\begin{enumerate}
\item Identify the extremal regions that represent the finger of a fingertip candidate. This applies to all parent regions of the candidate up to a maximum size and as long these do not have siblings, i.e. the candidate region is the only leaf contained in the parent's subtree. The largest of these regions will subsequently be referred to as \textit{finger}.
\item For each candidate region a feature vector is compiled containing the following properties:
\begin{itemize}
\item Bounding ellipse of candidate region (major and minor axis).
\item Bounding ellipse of finger region (major and minor axis).
\item The depth of the subtree having the finger region as its root node.
\item The maximum relative growth in pixel size of any child-parent relationship between the candidate and the finger region.
\item The relative growth in pixel size of the finger region with respect to the candidate region.
\item The ratio of the intensity ranges of the candidate and finger regions respectively.
\item The maximum value of the first of the seven shape descriptors introduced by \citeauthor{hu1962visual} for all regions below the finger region.
\item The number of pixels within a defined distance that have an intensity of less then $10\%$ of the candidate region's mean intensity. This feature is computed during this step for candidate regions only and is based on the observation that fingertips appear as spikes in the intensity height image (see figure \ref{fig:normalization}) hence resulting in a significant intensity fall-off in vicinity.
\end{itemize}
\item Calculate a confidence score $C$ using the weighted sum of the feature vector:
\begin{equation}
C = \sum\limits_{f_i}{w_i\cdot f_i}
\end{equation}
The weights have been chosen such that dimension differences between features are compensated without giving any of the features a too high influence on the final score.
\item Two thresholds have been defined on the confidence score to classify candidate regions either as \textit{no confidence}, \textit{low confidence} or \textit{high confidence}. However having a \textit{high confidence} score in a single frame is not sufficient for a region to be regarded as a fingertip. A candidate region must achieve in three consecutive frames at least once a \textit{high confidence} and never a \textit{no confidence} score in order to be classified as an actual fingertip.
\end{enumerate}

\subsection{Hand Distinction}\label{section:hand_distinction}
In order to establish the finger-hand relationships of identified contact points several approaches have been proposed in the literature. For instance \cite{dang2009hand} and \cite{wang2009detecting} use the orientation of the contact area's bounding ellipse to infer the relationship between hands and fingers. That feature however is highly dependent on the finger posture. While an \textit{oblique touch} yields a bounding ellipse with high eccentricity, hence the orientation can be inferred with high confidence, a \textit{vertical touch} in contrast results in a more circular shaped ellipse which does not provide useful orientation information \cite{wang2009detecting}. The assumption that the user always touches the surface at an oblique angle however requires prior knowledge by the user and might prove confusing for novice users. Another approach proposed by \citeauthor{dohse2008enhancing} is to use an overhead camera to group fingertips. Although that approach works reasonably well it requires an external camera hence thwarting the compactness of the setup while adding a significant additional processing overhead to the pipeline \cite{dohse2008enhancing}.

In this approach however the grouping can be robustly established using solely the information from the previous processing steps, that is the component tree and the identified contact points that to a given degree of confidence correspond to fingertips. These fingertips might or might not belong to the same hand or even the same user. As the respective extremal regions are leafs in the component tree they are contained in a number of regions of higher level until all of them are contained in the root region. Hence there already exists a spatial clustering of these regions based on intensity values. Therefore fingertips could simply be grouped gradually while ascending in the component tree until a homogeneity criterion such as the maximum distance between fingertips or the maximum size of the parent region is violated. While that approach generally works reasonably well problems arise under certain conditions. In the case of excessive ambient light such as from stray sun light the amount of contrast in the camera image is heavily reduced. However with reduced contrast less extremal regions are being revealed resulting in a sparse component tree. That loss of spatial information might lead to an erroneous clustering or a dead-lock situation when too many fingertips are child of the same region.

Therefore a more robust approach will be presented here that combines the advantages of the idea described above with agglomerative hierarchical clustering. Hierarchical clustering itself would be unsuited in our scenario because of its high computational complexity\footnote{Generally its computational complexity is bound by $\mathcal{O}(n^3)$, however in certain cases a bound of $\mathcal{O}(n^2)$ can be achieved\cite{defays1977efficient,sibson1973slink}}. Tabletop surfaces continually grow in size hence enabling the simultaneous interaction of multiple users. However due the large number of resulting touch points unaided nearest neighbor clustering would be highly inefficient for real-time application.

The idea is to only cluster a small subset of touch points at a time using agglomerative hierarchical clustering hereby reducing the impact of the polynomial complexity. The spatial clustering provided by the component tree, although erroneous at times, can serve as a reasonable input. The outline of the algorithm is as follows:

\begin{enumerate}
\item Traverse the component tree in postfix order. This ordering ensures that all children have been processed before the node itself is being processed. For each node do:
\begin{itemize}
\item If the node is a leaf node and has been classified with at least low confidence as fingertip, create a new cluster containing this node.
\item Otherwise create the current set of clusters from all child nodes and process these as follows:
\begin{enumerate}
\item Compute the distance matrix between all pairs of clusters of this node using a distance function $d(C_1, C_2)$. \label{clustering_start}
\item Find the pair of clusters ${C_1, C_2}$ with lowest distance $d_C(C_1, C_2)$ that still satisfies the distance criterion $\mathcal{D}(C_1, C_2, d_C)$.
\begin{itemize}
\item If such a pair of clusters exists, merge the two and continue with \ref{clustering_start}
\item Otherwise the clustering is finished.
\end{itemize}
\end{enumerate}
\end{itemize}
\end{enumerate}

Different distance functions such as \textit{single-link} (nearest neighbor) or \textit{complete-link} (furthest neighbor) exist. \textit{Single link} is defined by
\begin{equation}
d_C(C_1,C_2)=\min_{u\in C_1, v \in C_2}{d(u,v)}
\end{equation}
while \textit{complete-link} uses 
\begin{equation}
d_C(C_1,C_2)=\max_{u\in C_1, v \in C_2}{d(u,v)}
\end{equation}
with $d$ being a distance measure. Here Euclidean distance is used as distance measure.

\textit{Single link} would not be appropriate as a cluster could in certain cases grow too large hereby including fingertips from a different hand in proximity. \textit{Complete link} however considers the distance to the fingertip that is furthest away. Hence combining \textit{complete link} with an appropriate \textit{distance criterion} $\mathcal{D}$ that places a constraint on cluster growth has been found to work best to cluster fingertips into hands.

$\mathcal{D}$ is set to the maximum expected distance between any two fingertips of the same hand. The maximum distance is usually the one between the thumb and the little finger. However as user interaction using only these two fingers is rather rare the criterion has been restricted further based on the combined number of fingertips in the clusters. The lower the number of fingertips contained in both clusters the smaller is the maximum allowed distance. Hence fingertips that are at an unusually large distance from each other will only be merged once fingertips in intermediate positions have been found. Therefore this extension makes the criterion more robust as it alleviates negative effects of the chosen maximum hand distance being too large or too small.

\subsection{Hand and Fingertip Registration}
The registration of hands and fingertips in optical tabletop displays has recently received increased interest in the literature although the amount of research still remains limited. \citeauthor{walther2011left} use a decision tree to classify fingertip configurations and are currently the only ones to provide fingertip registration with less than five fingers being present. However their approach can only be considered a starting point for further research as the accuracy of around 80\% is still too low for productive use. One reason for the lack of accuracy is the anatomic variation of hands between humans. Although there are constraints on finger orientation and position, configurations differ highly based on hand and finger size.
Therefore current approaches require the presence of five fingers in order to perform fingertip registration. To meet that requirement one could imagine the presence of five fingers being the trigger to display a menu aligned around the hand as proposed in \cite{au2010multitouch} and \cite{micire2011hand}. The hand registration would for instance allow the display of different menus for the dominant and non-dominant hand.

\begin{figure}[tp]
	\centering
    \subfloat[Initial fingertip positions of a hand in resting position.]{\label{fig:registration_finger_1}\includegraphics[width=0.4\textwidth]{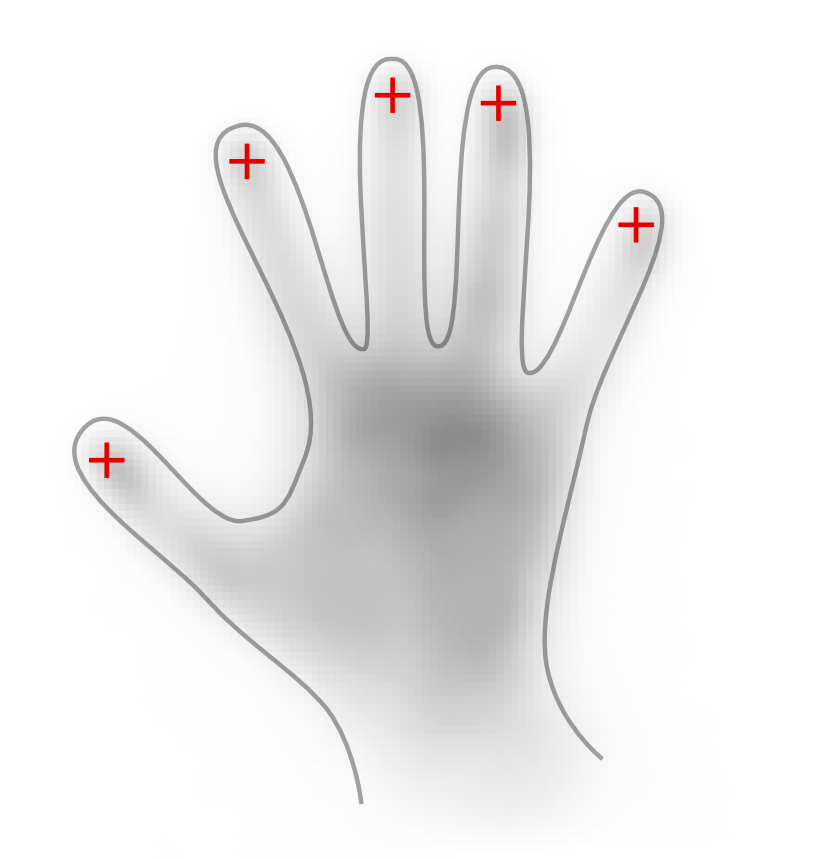}}
    \hspace{1cm}
    \subfloat[Fingertip ordering along shortest path.]{\label{fig:registration_finger_2}\includegraphics[width=0.4\textwidth]{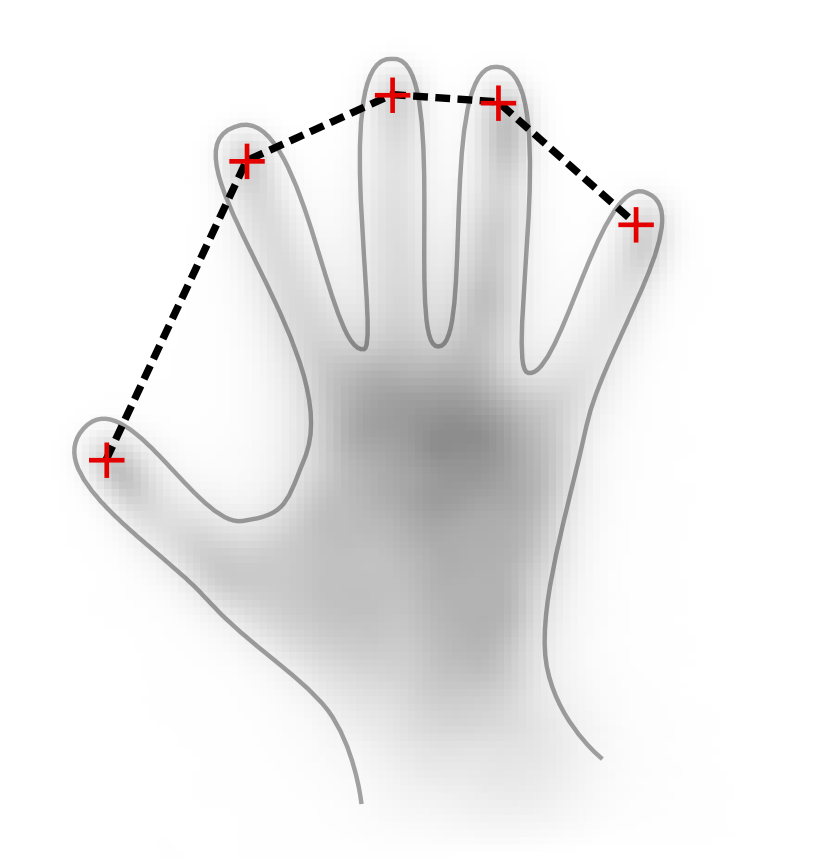}}\\
	\subfloat[Endpoints are known to be either thumb or little finger.]{\label{fig:registration_finger_3}\includegraphics[width=0.4\textwidth]{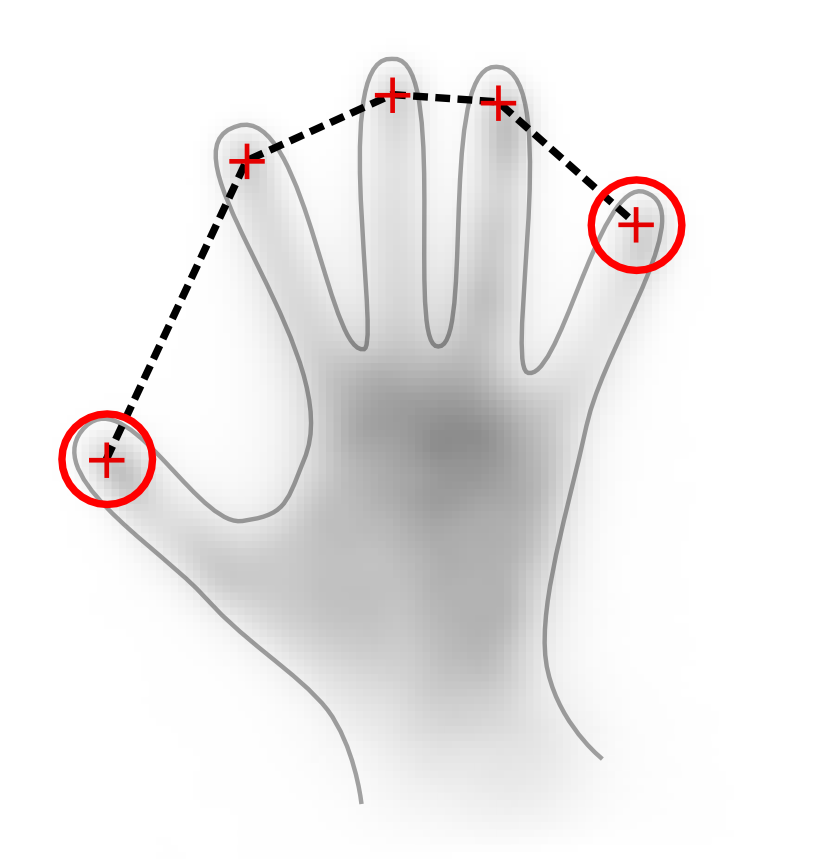}}
    \hspace{1cm}
	\subfloat[Identify thumb as the finger furthest away from the centroid of all five fingertips]{\label{fig:registration_finger_4}\includegraphics[width=0.4\textwidth]{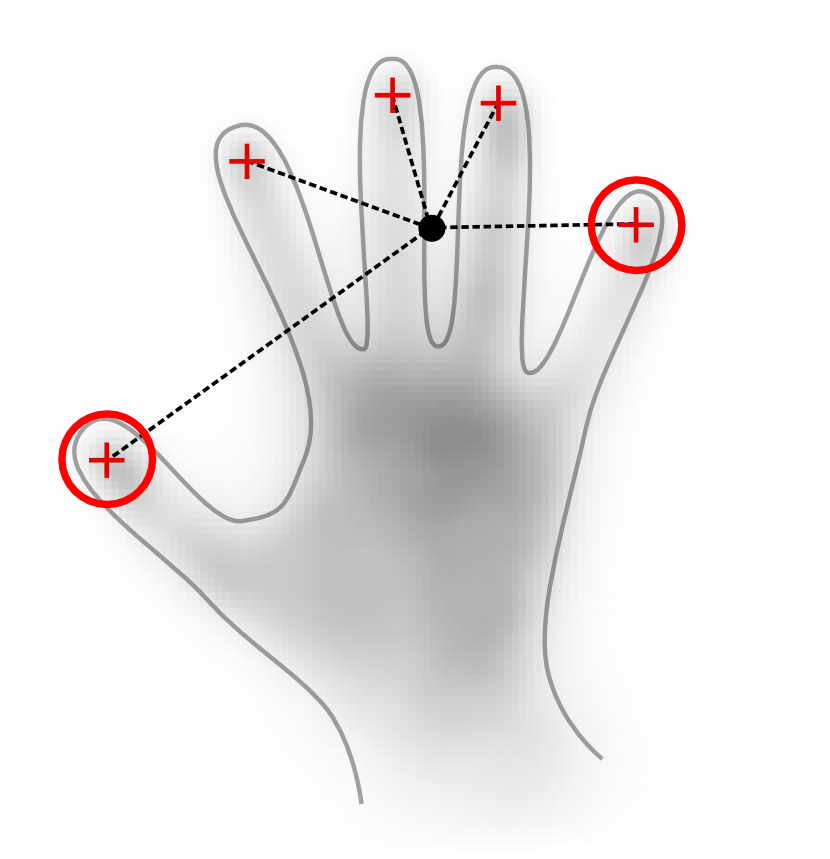}}\\
	\subfloat[Fingertips can now all be identified given the thumb (T) and the ordering of fingertips.]{\label{fig:registration_finger_5}\includegraphics[width=0.4\textwidth]{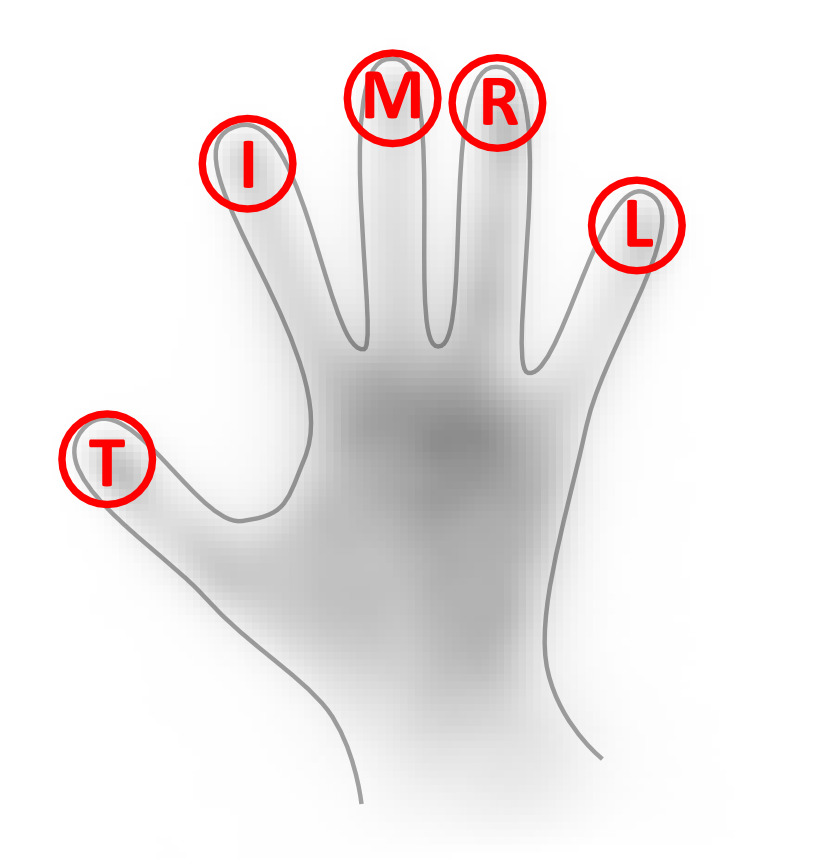}}

    \caption{Steps of the fingertip registration process. The hand shape and fingertip positions used in the drawing have been copied from an actual camera screenshot from the prototype.}
    \label{fig:registration_finger}
\end{figure}

Similarly to the above approaches the registration process comprises the following steps which are also visualized in figure \ref{fig:registration_finger}:
\begin{enumerate}
\item Order the five fingertips along the hand such that either the thumb or the little finger comes first.
\item Identify the thumb from the set of fingertips. The fingertip registration of the remaining fingertips follows from the previous ordering.
\item Infer from the fingertip registration whether the considered fingertips are from a left or right hand.
\end{enumerate}

The first step is crucial as an erroneous ordering can not be corrected at later stages of the registration process. \citeauthor{au2010multitouch} for instance propose an angular ordering around the centroid of the five fingertips \cite{au2010multitouch}. However the centroid usually deviates too much from the perceived center position as the index, middle, ring and little finger are generally located very closely to each other. While this approach is robust if fingertips are aligned as an arc on the surface, considering one of the fingers being in movement such as the index finger performing a sliding gesture downwards the ordering might swap thumb and index finger. Hence another ordering will be proposed here based on the shortest distance between fingertips. This ordering is similar to the one described in \cite{micire2011hand} which however was not known to the author at the time of elaboration as the mentioned paper was published only after the completion of the development of this prototype. The ordering is computed as follows:

\begin{enumerate}
\item Be $\mathcal{F}$ the set of all fingertips $f_i$. The set $\mathcal{D}$ is defined as the distances between all possible pairs of fingertips:
\begin{equation*}
\{\{d(f_i,f_j), f_i, f_j\}\;\mid\; (f_i,f_j)\in F\times F\}
\end{equation*}
with $d(f_i,f_j)$ denoting the Euclidean distance between $f_i$ and $f_j$.
\item For each fingertip $f_i$ assign a unique label $L(f_i)$.
\item Iterate $\mathcal{D}$ in increasing order of $d(f_i,f_j)$. Be $\mathcal{C}$ the set of all edges forming the fingertip contour. For each pair $(f_i, f_j)$ do
\begin{itemize}
\item If $L(f_i) \neq L(f_j)$:
\begin{enumerate}
\item Add $(f_i, f_j)$ to $\mathcal{C}$.
\item Assign a common unique label to all fingertips $f_k$ that either fulfill $L(f_k) = L(f_i)$ or $L(f_k) = L(f_j)$.
\end{enumerate}
\item If $\mid \mathcal{C} \mid = 4$:
\begin{enumerate}
\item All fingertips have been included in the contour. $(f_i, f_j)$ denotes the pair of fingertips on the contour that are furthest apart from each other. $f_i$ and $f_j$ define the endpoints of the contour which by definition of the ordering correspond either to the thumb or the little finger.
\end{enumerate}
\end{itemize}
\end{enumerate}

Obviously the above ordering relies on the assumption that on the contour thumb and little finger are the two fingertips that are furthest apart from each other. Since a rather unnatural hand pose would be required to violate that property it was deemed a valid assumption for regular use.

Therefore the next step in the registration process is to uniquely identify the thumb. As the previous ordering already reduced the number of candidate fingertips, the task is equivalent to distinguish thumb and little finger. Since the index, middle, ring and little finger influence the position of the centroid as mentioned above, thumb and little finger can be distinguished with respect to their distance to the centroid. The little finger is the one located closer to the centroid while the thumb is the one positioned further away.

\begin{figure}[tp]
	\centering
    \subfloat[]{\label{fig:registration_hand_1}\includegraphics[width=0.4\textwidth]{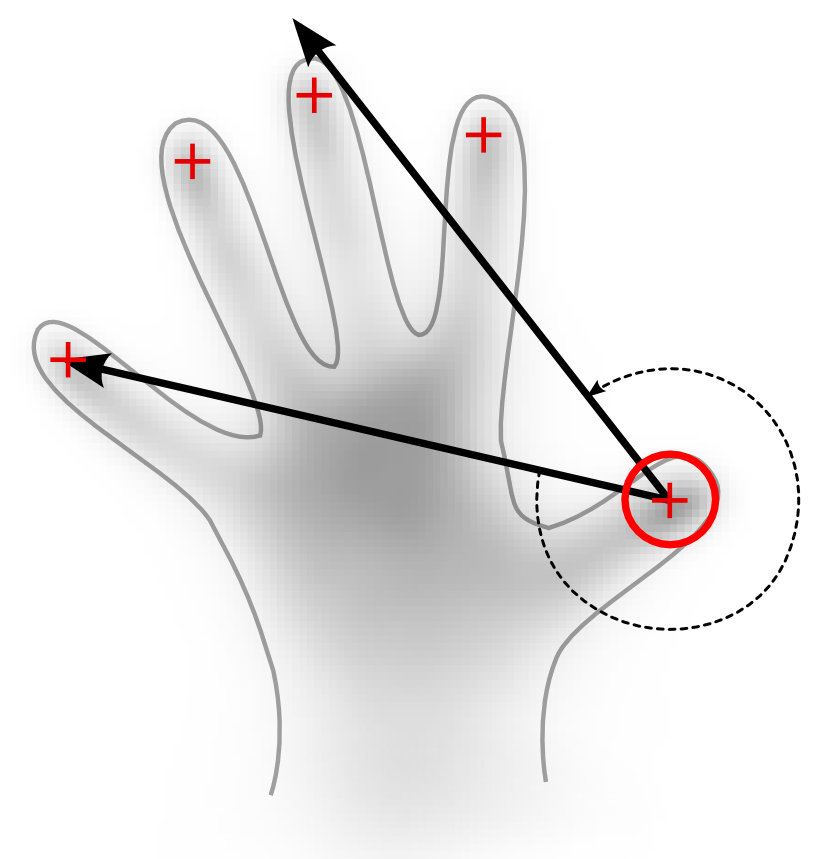}}
    \hspace{1cm}
    \subfloat[]{\label{fig:registration_hand_2}\includegraphics[width=0.4\textwidth]{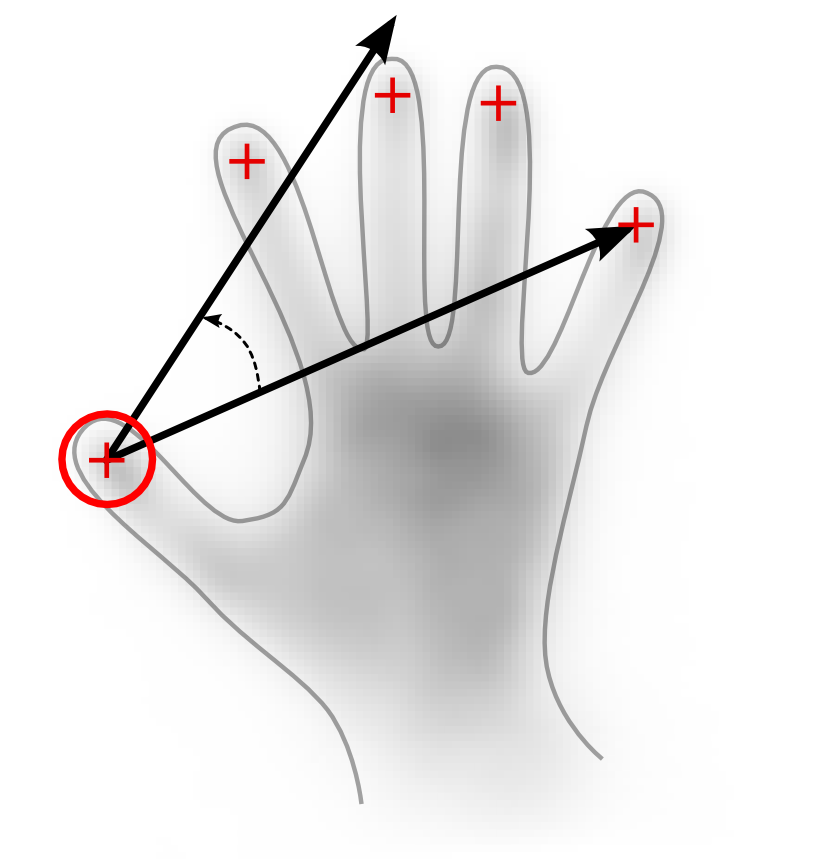}}
    \caption{Hand classification using the angle between little finger and the remaining fingertips with respect to the thumb.}
    \label{fig:registration_hand}
\end{figure}

Finally the classification of the set of fingertips as belonging to the left or right hand remains to be done (see figure \ref{fig:registration_hand}). Be $f_T, f_I, f_M, f_R$ and $f_L$ the thumb, index, middle, ring and little finger respectively. Based on the angle between the the vectors $\overrightarrow{f_T f_L}$ and $\overrightarrow{f_T f_I}$, left and right hand can be easily distinguished. If that angle is smaller than $180^\circ$ the set of fingertips is classified as right hand, otherwise as left hand. In order to make the classification more robust in presence of finger movement such as the aforementioned sliding down gesture of the index finger, the vector $\overrightarrow{f_T f_{IMR}} = \overrightarrow{f_T f_I} + \overrightarrow{f_T f_M} + \overrightarrow{f_T f_R}$ will be used instead of $\overrightarrow{f_T f_I}$ only.

Hence the classification is defined as follows:
\begin{equation}
\begin{array}{ccccl}
\mid\overrightarrow{f_T f_{L}} \times \overrightarrow{f_T f_{IMR}}\mid &=& \mid\overrightarrow{f_T f_{L}}\mid\cdot\mid\overrightarrow{f_T f_{IMR}}\mid\cdot\sin{\theta} &<& 0 \;\Rightarrow\; \mbox{Left Hand} \\
\mid\overrightarrow{f_T f_{L}} \times \overrightarrow{f_T f_{IMR}}\mid &=& \mid\overrightarrow{f_T f_{L}}\mid\cdot\mid\overrightarrow{f_T f_{IMR}}\mid\cdot\sin{\theta} &>& 0 \;\Rightarrow\; \mbox{Right Hand}
\end{array}
\end{equation}
where $\times$ denotes the cross product of two vectors.

\subsection{Tracking}
The tracking step aims to establish intra-frame correspondences of the detected fingertips and similarly to the default processing pipeline a two step approach is chosen here. First for all previously detected fingertips the estimated position is predicted in the current frame in order to simplify matching. However instead of the Kalman filter which is computationally demanding the much faster and equally accurate \textit{double exponential smoothing} algorithm will be used as a motion model. Second using the predicted positions of previous blobs in the current frame, relationships are established using nearest neighbor matching. However given the previous clustering of fingertips this step can be significantly accelerated. Instead of considering all possible combinations of fingertips between the two frames, this number can be narrowed down by only considering combinations of fingertips whose containing cluster intersect. In order to test for an intersection the bounding boxes of the two clusters have been used.

In the following section the \textit{double exponential smoothing} algorithm will be outlined. However as the following nearest neighbor matching is basically the same as in the default processing pipeline, the reader is referred to page \pageref{section:default_pipline:tracking} for an in-depth description.

\paragraph{Double Exponential Smoothing} \hfill\\

The \textit{double exponential smoothing} algorithm as proposed by \citeauthor{laviola2003double} in \cite{laviola2003double} is based on the assumption that movements can be accurately approximated using linear equations. Obviously this only holds true for short prediction intervals. However as multi-touch applications require a high update frequency for the sake of interactivity, this does not appear to be an actual constraint. Given this assumption a simple yet performant algorithm design can be derived.

For the approximation of the y-intercept and the slope of the linear movement equation two smoothing statistics are defined. These are calculated based on an exponential weighting of movement positions giving recent positions a higher weight compared to older ones. The degree of exponential decay is determined by a parameter $\alpha\in[0,1]$. Be $\vec{p}_t$ the position vector at time $t$, therefore the \textit{smoothing statistics} $\vec{S}_{\vec{p}_{t}}$ and $\vec{S}^{[2]}_{\vec{p}_{t}}$ are calculated as follows:
\begin{align}
 \vec{S}_{\vec{p}_{t}} = \alpha\vec{p}_t + (1-\alpha)\vec{S}_{\vec{p}_{t-1}}\label{double_exp_equ_spt}\\
 \vec{S}^{[2]}_{\vec{p}_{t}} = \alpha\vec{S}_{\vec{p}_t} + (1-\alpha)\vec{S}^{[2]}_{\vec{p}_{t-1}}\label{double_exp_equ_spt2}
\end{align}

Equation \ref{double_exp_equ_spt} calculates an exponentially weighted position value based on previous positions. This calculation is then again smoothed exponentially in equation \ref{double_exp_equ_spt2}. Based on those two values an approximation of the linear movement equation can be derived allowing a position prediction at time $t+\lambda$. For a detailed algebraic calculation the interested reader is referred to \cite{laviola2003double}. It follows:
\begin{equation}
 \vec{p}_{t+\lambda} = \left(2 + \frac{\alpha}{1-\alpha}\cdot\lambda\right)\cdot \vec{S}_{\vec{p}_t} - \left(1 + \frac{\alpha}{1-\alpha}\cdot\lambda\right)\cdot \vec{S}^{[2]}_{\vec{p}_t}
\end{equation}

In order to achieve a correct prediction the algorithm assumes that movement positions are captured at equal intervals. Hence the algorithm is only able to predict at times $\lambda=k\cdot\tau$ with $k\in\mathbb{N}$. 
Since predictions are only required for camera frames which are delivered at a fixed rate, this is no constraint in our usage scenario.

In \cite{laviola2003double} the algorithm has been compared to the widely used Kalman filter in terms of efficiency and accuracy. It showed that the prediction of human movements was possible with almost equal accuracy while outperforming the Kalman filter by factor 135.

\subsection{Multi-Touch Abstraction Layer}
Generally it is desirable to separate the multi touch sensing technology and the actual multi touch application using a simple abstraction layer. Although this introduces additional latency to the processing pipeline it would allow the distribution of computation across a network or the usage of a different programming framework for application development. The latter is especially important as application developers from a design background often use specialized high-level frameworks such as \textit{Processing}, \textit{Pure Data} or \textit{Flash}. The TUIO protocol is a network-based abstraction layer originating from the reacTIVision toolkit that has been widely adopted in the multi touch research and open source community since its initial publication in 2005 \cite{kaltenbrunner2009reactivision}. 
\paragraph{TUIO}
The TUIO protocol has been developed to provide a simple way of distributing touch states from the reacTable without introducing significant additional latency. Although it was tailored to suit the needs of the original project it got widely adopted because it supported the basic primitives, i.e. touch points (called \textit{cursors}) and markers (called \textit{objects}. The specification of markers will not be included here, see \cite{kaltenbrunner2005tuio} for a detailed description). In order to achieve a low-latency transmission the following characteristics are notable:
\begin{itemize}
\item TUIO messages are based on Open Sound Control (OSC) a syntax widely used to encode control messages for musical instruments \cite{kaltenbrunner2009reactivision}. OSC messages are defined as a tuple containing the message profile, the command, the argument type signature and the argument array. Several messages can be included in a bundle to reduce the number of transmitted packets.
\item TUIO messages/bundles are transmitted over the network using UDP. UDP provides fast transmission at the lack of state information, i.e. the sender is unaware whether the packet actually arrived at the destination.
\item In order to ensure proper functioning using an error-prone transmission channel, TUIO communication is based on touch states instead of touch events.
\end{itemize}
The TUIO update message for \textit{cursors} is defined as
\begin{center}
\begin{tabular}[t]{l}
\texttt{/tuio/2Dcur set sid xpos ypos xvel yvel maccel}
\end{tabular}
\end{center}
containing as arguments the session id, position, velocity and acceleration. The profile also exists for the 2.5D (containing the height above the surface) and 3D (containing 3D position and rotation information) cases, see \cite{kaltenbrunner2005tuio} for a detailed description.

The TUIO protocol furthermore includes the alive states for \textit{cursors}:
\begin{center}
\begin{tabular}[t]{l}
\texttt{/tuio/2Dcur alive [active session IDs]}\\
\end{tabular}
\end{center}
The session IDs of all active \textit{cursors} must be included in this messages during their whole lifetime regardless of whether their state has been previously updated using the aforementioned messages.

In case messages do not arrive in orderly fashion at the destination, messages can be reordered or dropped using the additional frame sequence id:
\begin{center}
\begin{tabular}[t]{l}
\texttt{/tuio/2Dcur fseq [int32]}\\
\end{tabular}
\end{center}

\paragraph{Custom TUIO profiles}
Although the TUIO protocol provides profiles to efficiently represent single touch points it lacks support for hands. However the protocol allows the inclusion of custom commands that will be considered if supported by the receiving application or simply ignored otherwise. Hence the following profiles are to be considered complementary to the above described standard set of messages.

The custom hand profile defines two messages:
\begin{center}
\begin{tabular}[t]{l}
\texttt{/custom/\textunderscore{}hand set sid type posx posy width height [array of 2Dcur sids]}\\
\texttt{[array of finger types in the same order as previous sids]}\\
\newline\\
\texttt{/custom/\textunderscore{}hand alive [active session IDs]}\\
\end{tabular}
\end{center}
The former updates the hand parameters, that is the hand \texttt{type} ("left", "right", "unknown"), its bounding box ((\texttt{posx}, \texttt{posy}) defines the center of a rectangle with edge lengths \texttt{width} and \texttt{height} respectively) and the touch points associated with this hand. Additionally for each touch point the finger registration is included (t - thumb, i - index finger, m - middle finger, r - ring finger, l - little finger, u - unknown). The latter message works similarly to the \texttt{alive} messages defined above.

\section{Implementation}
During the work on the tabletop prototype an image processing framework has been developed that implements the proposed approach to multi-touch processing. The framework has a number of notable characteristics:
\begin{description}
\item[Performance] Performance might seem like an obvious requirement for real-time applications however usually poses a real challenge. At best its computational activity goes unnoticed to the user and multi-touch interaction is highly responsive. In the worst case it introduces latencies which interrupt user interaction at a level to making it unusable. Hence the framework was designed to make the most of todays multi-core CPUs and therefore relies highly on multi-threading wherever possible and appropriate. However threading also requires synchronization which in turn results in a computational overhead. In image processing frameworks such as OpenCV synchronization usually takes place after each image processing operation. However given the number of processing steps in the proposed pipeline this might add up to a significant overhead. Thus image processing operations are bundled into smaller processing pipelines that will be executed without synchronization. 
\item[Interoperability]
In order to being able to use this framework with other applications, also those that have not explicitly been developed for this framework, the TUIO protocol has been chosen as a means of communication. The TUIO protocol has established itself as a de facto standard in the open-source and most parts of the research community. The framework implements the TUIO v1.1 specification\footnote{see \url{http://www.tuio.org/?specification}} that is understood by most current applications hence allows the communication of fingertip positions. However the proposed extended commands which are transmitted alongside the common command set will only be interpreted by those applications having been developed for this framework.
\item[Extensibility]
The framework has been developed as a proof-of-concept of the proposed approach, hence the implemented image processing methods mainly reflect those described previously. However all image processing operations are based on a common base class that enables their easy integration in the processing pipeline. Thus, the processing pipeline can be extended to include any number of additional functionality without any modifications to the core components.
\item[Platform Independence]
Although this framework has been solely developed on a machine running Ubuntu\footnote{see \url{www.ubuntu.com}} Linux in version 10.04, platform independence is an important criteria to ensure adoption of the framework by a wider public. Hence all accesses to hardware or operating system (OS) resources have been wrapped by dedicated libraries that ensure proper operation on a variety of hardware and operating system configurations. For instance, the boost\footnote{see \url{http://www.boost.org}} libraries have been used for access to the file system and threading capabilities of the host OS. Furthermore for accessing cameras based on the ieee1394\footnote{see \url{http://www.1394ta.org/} and \url{http://damien.douxchamps.net/ieee1394/libdc1394/iidc/} for the actual specifications} standard the dc1394\footnote{see \url{http://damien.douxchamps.net/ieee1394/libdc1394/}} library has been employed, while the widely used computer vision library OpenCV\footnote{see \url{http://opencv.willowgarage.com/}} has been used to implement some image processing features.
\end{description}
Additional credits go to Daniel Fischer who developed a similar framework for the previous prototype at the Virtual Reality Systems Group. The provided source code for camera frame acquisition, image rectification and a control interface based on OpenGL have been thankfully reused in this framework. 

\subsection{Image Processing}
Image processing has been implemented similarly to other frameworks with one notable exception. Unlike other frameworks, where image operators usually process the entire image, all image processing functionality has been implemented to being able to operate on both, the entire image or just a smaller, predefined area of the image. This is especially useful with operations which apply special processing to border cases. For instance the connected components algorithm described on page \pageref{section_connected_components_horn} uses previously computed values to determine the value of the current pixel. This however requires pixels on the image border to be treated differently from pixels inside the image increasing the number of checks during operation. For all pixels contained in the first row and the first column of the image these previous values just do not exist as they lie outside the image area. However to streamline the algorithm implementation and to reduce the number of required checks, the algorithm processes an extended image which has a one pixel border initialized with a default value added around the actual image data. While traditional implementations would require copying image data to the extended image at each frame, results of previous operations can be written directly into the extended image using the capabilities of this framework. As the border pixels are only required for reading and are never written, these only need to be initialized once at startup.

\subsection{Multi-Threading}
As has been previously shown in figure \ref{fig:pipeline} on page \pageref{fig:pipeline}, the processing pipeline provides the possibility of concurrent execution during two stages:
\begin{description}
\item[Preprocessing] As all preprocessing steps starting from \textit{distortion correction} until \textit{region of interest detection} operate on a per-pixel basis, they can be considered a single combined operator when it comes to threading. Thus the image processing area is partitioned between the number of available threads and the whole preprocessing pipeline is executed entirely without synchronization.
\item[Region of Interest Processing] After the detection of regions of interest in the image the threading strategy changes. Regions of interest are each processed in a single thread while the processing of several regions of interest is distributed among all available threads in decreasing region size. Hence the impact of threading depends on the number of regions of interest and their size. 
\end{description}
The framework provides specialized classes that allow the easy implementation of this functionality. By chaining a number of image operations in \textit{ProcessSets} a processing pipeline can be defined that runs on a single thread. However \textit{ProcessSets} can be grouped again and be marked for parallel execution enabling the execution of completely different pipelines on different threads. In order to provide simple means for synchronization execution barriers can be defined until either all or a predefined \textit{ProcessSet} has finished.

%% file: Content/Chapter_3.tex
\chapter{Evaluation}\label{chapter:evaluation}
In this chapter the previously presented processing pipeline is being tested in terms of performance and detection accuracy. Performance is obviously a crucial aspect of a real-time application as high latencies will be immediately felt by the user and impact usability. However detection accuracy might be considered even more important due to the high disruptive potential that erroneously identified touches might have on user interaction.

Given the importance of these two factors, the presented evaluation aims to test the limits of the proposed processing pipeline by including as many simultaneous users as possible in the interaction. As to the maximum number of people that might comfortably perform simultaneous two-handed interaction on the multi-touch prototype without getting too much in their way, four people each of them standing on one side of the table has been found to work best. Hence, four students (all male, aged 24 - 27) from the Virtual Reality Systems Group at the Bauhaus University Weimar were asked to take part in the evaluation, resulting in a maximum number of 40 simultaneous finger touches (and possibly even more evaluated touch points considering that users were allowed to rest their palm on the surface).

Since no multi-touch capable user interface had been developed at this point, the participants were asked to mimic multi-touch gestures on the blank surface. However they were reminded to not only use common gestures involving a small number of simultaneous touches but to also use both hands with randomly changing configurations of fingers touching the surface. Due to the lack of a real application participants tended to perform rather swift and abrupt movements though. While that is considered here as a challenge for the processing pipeline it obviously does not reflect real-world usage. We assume that given a calmer user interaction the measured accuracy might be even better.

In the end more than 4000 camera frames were captured during the interaction process that now could be replayed to the processing pipeline in order to properly measure execution times and detection accuracy. 

\section{Performance}
As performance is dependent on the number of simultaneous user actions performed on the surface, projects like the Microsoft Surface define a maximum threshold on simultaneous touches to avoid exceeding real-time performance. However as surfaces continuously grow larger and larger in size, processing pipelines have to deal with an ever increasing number of simultaneous touches. Therefore evaluating the detection performance for up to 40 simultaneous fingers touching the surface is being considered a well-suited performance indicator of the processing pipeline.

Measurements have been taken on a machine featuring a Intel Core i7 CPU940 (2.93GHz, 4 cores, 8 logical processors using Hyper-Threading\footnote{see \url{http://www.intel.com/content/www/us/en/architecture-and-technology/hyper-threading/hyper-threading-technology.html}}) and 5.8GB of memory running Ubuntu Linux in version 10.04. The camera delivered frames at a rate of 60 fps at a resolution of 640 x 480. The measured execution time includes all steps from the pipeline apart from image acquisition. Measuring was started just after the image had been loaded into memory and stopped when all processing and memory clean-up steps had been executed. In order to minimize the influence of external factors such as the operating system, measurements have been averaged over 5 processing runs of the captured user interaction.

\begin{figure}
	\begin{tikzpicture}[y=.5cm, x=.35cm,font=\sffamily]
		\draw (0,0) -- coordinate (x axis mid) (40,0);
			\draw (0,0) -- coordinate (y axis mid) (0,16);
			\foreach \x in {0,5,...,40}
		 		\draw (\x,1pt) -- (\x,-3pt)
				node[anchor=north] {\x};
			\foreach \y in {0,2,...,16}
		 		\draw (1pt,\y) -- (-3pt,\y) 
		 			node[anchor=east] {\y}; 
		\node[below=0.8cm] at (x axis mid) {Evaluated Fingertips};
		\node[rotate=90, above=0.8cm] at (y axis mid) {Execution Time (ms)};

		\draw plot[mark=*, mark options={color=orange,ultra thick}] 
			file {Data/performance_1.csv};
		\draw plot[mark=square*, mark options={color=magenta,ultra thick} ] 
			file {Data/performance_2.csv};
		\draw plot[mark=pentagon*, mark options={color=cyan,ultra thick}]
			file {Data/performance_4.csv};
	
		\begin{scope}[shift={(30,2)}] 
		\draw (0,0) -- 
			plot[mark=*, mark options={color=orange,ultra thick}] (0.25,0) -- (0.5,0) 
			node[right]{1 Thread};
		\draw[yshift=\baselineskip] (0,0) -- 
			plot[mark=square*, mark options={color=magenta,ultra thick}] (0.25,0) -- (0.5,0)
			node[right]{2 Threads};
		\draw[yshift=2\baselineskip] (0,0) -- 
			plot[mark=pentagon*, mark options={color=cyan,ultra thick}] (0.25,0) -- (0.5,0)
			node[right]{4 Threads};
		\end{scope}
	\end{tikzpicture}
\caption{Execution times with respect to the simultaneous number of finger touches for different numbers of threads involved in the processing.}
\label{fig:evaluation:performance}
\end{figure}
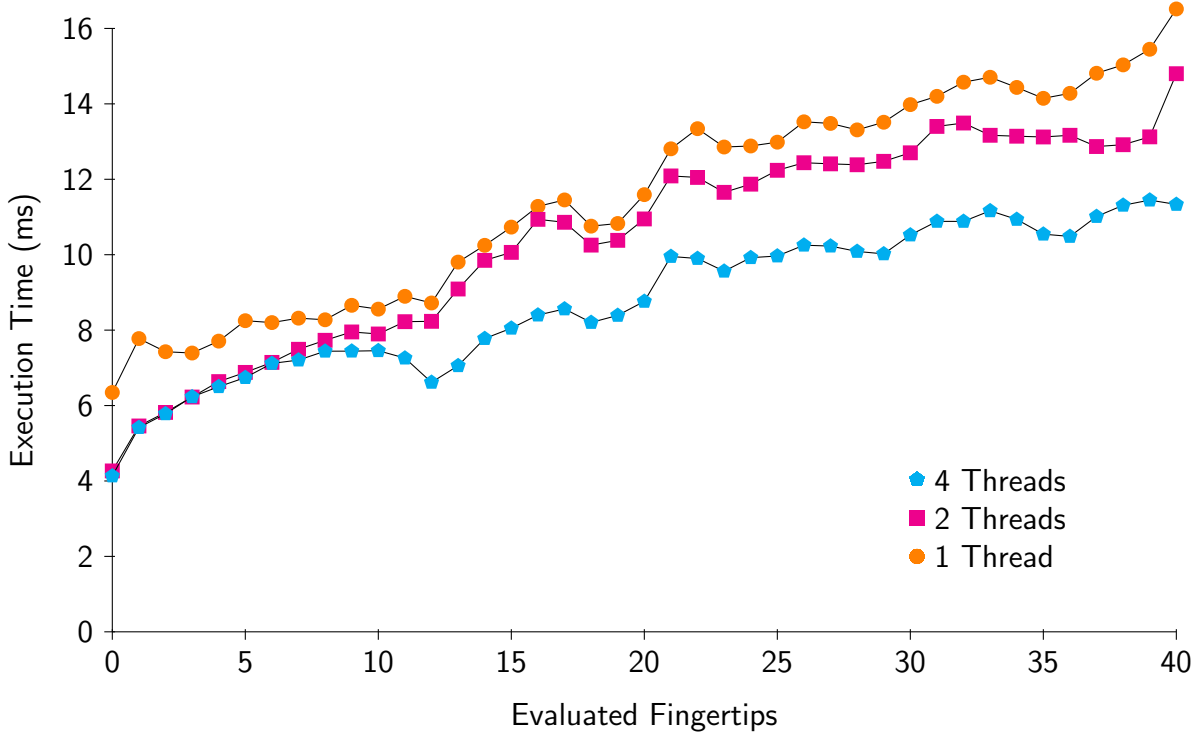

Given the measurements shown in figure \ref{fig:evaluation:performance} the most important result is that even for single threaded execution a processing performance of at least 60 frames per second is achieved. Most notably that performance was achieved even during the simultaneous interaction of 4 users, 8 hands and up to 40 fingers. Furthermore using four processing threads processing time is capped at around 10ms regardless of how many fingers of the 8 hands simultaneously touch the surface.
 
From the measurements it becomes obvious that performance is highly dependent on the way users interact with the surface. While 4 users each touching the surface with one finger would benefit from threading in the processing stage, one user placing four fingers of the same hand onto the tabletop would not be able to benefit from threading. Furthermore processing performance is determined by the largest and most complex (in terms of extremal regions) region of interest. Although the processing pipeline assigns regions of interest to threads in decreasing size, a region of interest that deviates too much in terms of size and complexity could in the end negatively impact performance as other threads would be left in idle state waiting for that region to be processed.

In the interaction process on which this evaluation is based, participants were asked at the beginning to place their hands on the surface one after another. At a later time of the interaction process rarely less than 10 simultaneous touches occurred and usually all 8 hands remained above the surface, hence measurements for less than 10 fingertips are largely based on frames where only one or two regions of interest were visible. Thus the impact of threading is limited in this case. Hence the performance progression with respect to fingertips should only be seen as an indicator of how performance could evolve.

Albeit the performance measurements being very satisfying it is only a part of the processing that is performed between a user action and the corresponding visual feedback. However it is that latency that in the end determines the users impression of responsiveness of a prototype. The contributing factors to the latency are as follows:
\begin{itemize}
\item Since internal camera operation time is negligible the major amount of latency results from the exposure time as the camera image is not created at a single instant in time but rather results from the integration of incoming light over a period of time.
\item Next the transmission of image data from the camera to the computer adds further latency. The camera used in the prototype transfers data at 400Mb/s. However IEEE 1394 devices send data packets at a fixed cycle rate hence it is actually the number of packets that determines the transmission time. According to the IIDC specification\footnote{IIDC (Instrumentation and Industrial Control Working Group) is the working group responsible for the specification. The specification is sometimes also known as Digital Camera (DCAM) specification. See \url{http://damien.douxchamps.net/ieee1394/libdc1394/iidc/}} the image mode used in this prototype (resolution: 640x480, pixel bit depth: 8, frame rate: 60) corresponds to a packet size of 2560 pixels. Thus $640*480/2560=120$ packets need to be send in order to transmit a complete camera frame. The camera has a cycle rate of $125\mu s$ hence transmission time adds another $120 * 125\mu s = 15ms$ to the latency. 
\item The processing pipeline is the next important contributor to latency as can be seen from the measurements. Furthermore latency can temporarily vary due to unfavorable process scheduling from the operating system.
\item The TUIO protocol used to communicate detection results adds further latency due to the required network transport. Although it relies on UDP to transfer data, which unlike TCP operates connection-less and therefore is faster, it still adds some milliseconds to the latency.
\item Finally the multi-touch application has received user input from the touchscreen although it still needs to compute a response to the input and update the user interface accordingly. Although this part is at the responsibility of the application developer it still needs to be taken into account that a major chunk of processing is still to follow.
\end{itemize}

Hence the latency introduced by the multi-touch sensing technology of the prototype can be summed up as follows:
\begin{itemize}
\item Camera Image Acquisition: 16ms
\item Camera Image Transfer: 15ms
\item Processing Pipeline: 10ms (with 4 threads)
\end{itemize}
The impact of the exposure time could be further reduced by pulsing the infrared light. Since the light's intensity would be much higher during the pulse the exposure time could be significantly shorter. While this would furthermore reduce the influence of ambient illumination, it would require additional logic to synchronize the illuminators and the camera. 
Moreover the camera transfer rate could be easily halved by using an IEEE 1394 compliant camera that supports the faster data transfer rate of 800Mb/s.
Given these possibilities to further shorten the sensing latency with additional hardware, the processing pipeline seems to be well suited performance-wise for applications in real-time scenarios.

\section{Accuracy}
The evaluation of accuracy is especially important due to the high disruptive potential that erroneous sensing results might have on usability. On the one hand wrongly identified touches might lead to unintended application behavior leaving the user puzzled about what he has done wrong. On the other hand missing an intentional touch can lead to increased user frustration as the user might need to retry several times to perform an action.

However measuring accuracy is particularly difficult as one first needs to establish a ground truth that later serves as a basis for comparison. In this case the first 1500 frames from the interaction process have been considered. Given the extensive manual work required to label all visible fingertips in these camera images, this number has been further reduced to only consider every fifth frame. Hence a total of 300 camera images have been analyzed that contained on average more than 21 visible fingertips.

In order to quantify the accuracy two basic measures will be used here. First will be considered how many of the visible fingertips have been correctly identified by the pipeline. This measure is usually called the \textit{hit rate}. Secondly, given the total number of identified fingertips, how many of these have been wrongly considered to be in contact with the surface. This second measure is usually referred to as \textit{false positive} rate. The measured accuracy is as follows:
\begin{description}
\item[Hit Rate]
$$
\frac{\mbox{Correctly identified fingertips}}{\mbox{Total number of visible fingertips}} = \frac{6449}{6628} \approx 0.973
$$
\item[False Positive Rate]
$$
\frac{\mbox{Wrongly identified fingertips}}{\mbox{Total number of identified fingertips}} = \frac{88}{6537} \approx 0.0135
$$
\end{description}
While the \textit{hit rate} is very satisfactory and shows the potential of the proposed pipeline it is the \textit{false positive} rate that might still need improvement. Given the constraints from the prototype in particular the significantly uneven illumination, this value appears to be still reasonably low though. Hence these two measures serve as an indication of the accuracy of this prototype but would require a re-evaluation in the future since the current data is based on a somewhat artificial user interaction as no proper user interface was present at the time.

Moreover the accuracy of the hand distinction has been analyzed as well. For this purpose more than 450 camera images have been manually classified which have been extracted from the interaction process at five frame intervals. The evaluation measured the precision of the hand distinction process, that is the percentage of hands that have been clustered correctly. Hand distinction was regarded as successful if all detected fingertips from the same hand were attributed to the same cluster. However if these fingertips were contained in more than one cluster the clustering for this hand was considered invalid. Furthermore if even two hands were erroneously contained in the same cluster both were regarded as unsuccessful. The measured accuracy is as follows:
\begin{description}
\item[Precision]
$$
\frac{\mbox{Correctly clustered hands}}{\mbox{Total number of visible hands}} = \frac{2681}{2909} \approx 0.922
$$
\end{description}
The hand distinction usually failed in the presence of \textit{false positives} from the fingertip detection. Since cluster size was capped at five fingers, \textit{false positives} resulted in surpassing that limit and hence interrupted the merging of clusters. Thus reducing the \textit{false positive} rate of the fingertip detection step would have positive effects on hand distinction in terms of \textit{precision}.

The accuracy of the hand and fingertip classification has only informally been tested. In total eight students, all male aged in their mid-twenties, from the Virtual Reality Systems Group at the Bauhaus University Weimar had been asked to place both of their hands on the surface. While no restrictions were imposed on the exact hand posture, the students were asked to adopt a position that felt natural to them. In these short tests the hand and fingertips have been registered correctly for all participants. Hence the assumptions used to infer these properties can be considered justified. Although only a static hand position was tested here, it currently seems to be the most convincing use case where a gesture is started in this position allowing a proper registration.

%% file: Content/Chapter_4.tex
\chapter{Conclusion}\label{chapter:conclusion}
In this thesis a novel processing pipeline for optical multi-touch surfaces has been presented motivated by the perceived lack of viable processing alternatives to enable real multi-hand and multi-user interaction. While the common approach usually relies on thresholding to discern surface contacts, this novel approach is centered around the concept of extremal regions which in contrast only describe a relative relationship between an image region and its border. Depending solely on this relative relationship these regions are much more robust in the presence of non-uniform illumination. Furthermore extremal regions can be organized in a hierarchical structure hereby revealing the spatial relationship between surface contacts. In order to efficiently compute these structures the processing pipeline relies on the Maximally Stable Extremal Regions algorithm. Although it has been successfully used in stereo image matching and visual tracking, this algorithm has never before been employed for multi-touch sensing. Given that hierarchical structure of extremal regions, a method has been described to reveal all those regions that correspond to a touch by a fingertip based on a set of image features. Based on these fingertips a second algorithm enables the attribution of fingertips to different hands based on a approach combining the hierarchical structure as well as agglomerative hierarchical clustering. Finally a method has been presented that provides hand and fingertip registration for these clusters of fingertips.

Subsequently the novel processing pipeline has been evaluated measuring both the performance as well as the detection accuracy. In total the processing of over 4000 frames has been taken into account to measure the pipeline's performance. Furthermore over 300 frames have been manually analyzed and labeled, hereby establishing a ground truth that served as a basis for comparison in the accuracy evaluation. The performance evaluation has shown that the required processing achieves real-time performance even with multiple user simultaneously interacting with both their hands. Moreover the accuracy evaluation provided very satisfying results bearing in mind that this is only a first prototype of the novel processing pipeline.

\section{Future Work}
This section presents a short outlook on potential directions for further development. These include work on the prototype itself as well as applications that make use of the newly introduced features.
\subsection{Machine Learning supported Fingertip Detection}
Although the accuracy results from the evaluation are very promising, the fingertip detection procedure might require further development. Considering both the \textit{hit rate} and \textit{false positive} rate it is evident that the procedure reveals fingertips with a very high confidence. However given the \textit{false positive} rate of $1.35\%$ there is still room for improvement. As the detection is already based on a numerical feature vector, one possible solution might be to use machine learning techniques like support vector machines for classification of extremal regions as fingertip or not. Since most of these algorithms however require a training data set of labeled examples of true and false fingertips, this comes at the cost of significantly higher manual work.

\subsection{Marker Support}
\begin{figure}[t]
    \includegraphics[width=\textwidth]{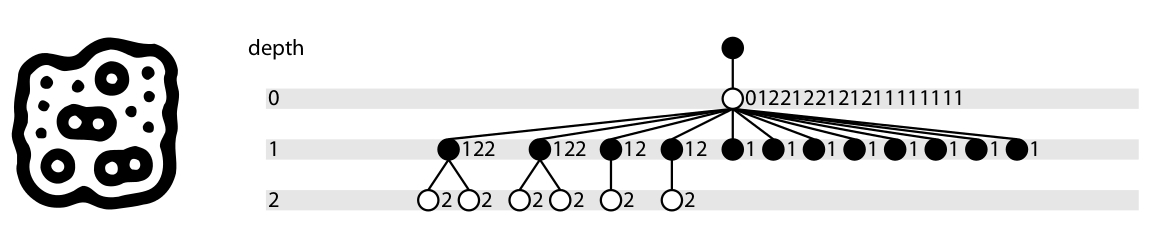}
    \caption{Amoeba fiducial and its topological structure \cite{bencina2005improved}}
    \label{fig:amoeba_fiducial}
\end{figure}

While the user's hands are usually the main input modality on tabletop displays they also lead to the user interface being the sole way of conveying information and functionality. However in certain situations tangible input devices that can be placed upon the tabletop might make functionality easier to grasp and hence might be considered a welcome extension of tabletop interfaces. The reacTable is a widely known prototype that intuitively combines both tangible objects as well as traditional multi-touch as user interaction techniques. Quoting the reacTable here is no coincidence as their marker engine, the Amoeba Engine, bears certain similarities to the processing used in this pipeline. Amoeba fiducials are characterized by the topological structure of black and white regions in the marker as shown in figure \ref{fig:amoeba_fiducial}. Enforcing additional constraints on the tree structure makes marker detection in the Amoeba engine very robust. Hence one could imagine designing a similar marker engine based on the component tree that is already computed in our processing pipeline. As the component tree only includes regions that are brighter than their surrounding their fiducials would obviously not be compatible with such an approach. However since amoeba fiducials only require a tree depth of 2, three gray levels (preferably white, gray, black to achieve maximum contrast) would be sufficient to build a comparable tree structure. Though it remains to be shown whether such a structure could achieve a similar robustness without the use of additional features.

\subsection{User Interaction Studies}
The novel processing pipeline and algorithms have been shown to reliably detect multi-touch user input while revealing further input properties such as the hand used for interaction. However all this is only a means to an end, that is enabling novel user interaction techniques. While the pipeline has been evaluated on a technical level, the important next step would be to make use of these features in multi-touch applications and study whether these properties can make interaction on tabletop displays a more seamless and intuitive experience.

\begin{figure}[t]
    \includegraphics[width=\textwidth]{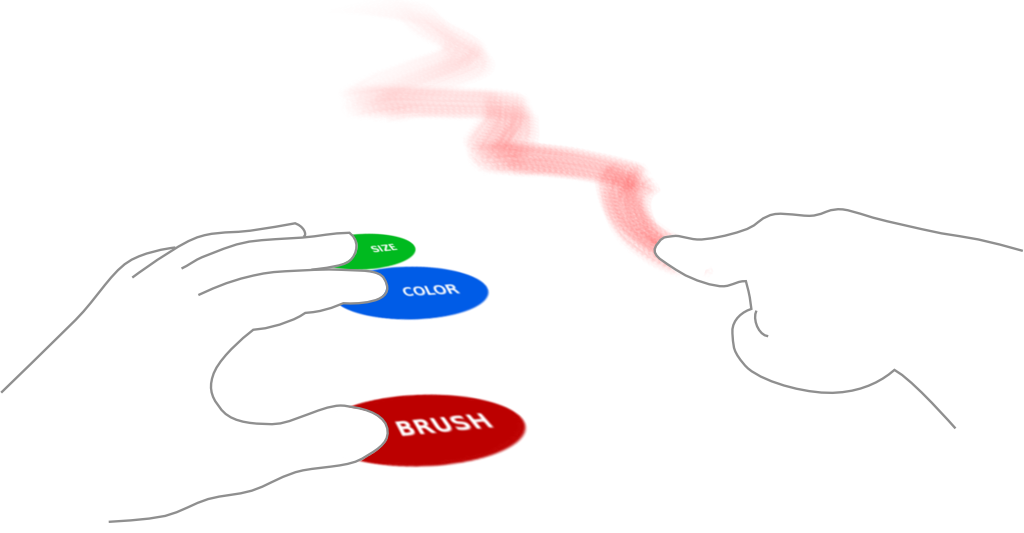}
    \caption{Design mock-up of a multi-touch application taking advantage of the additional interaction features.}
    \label{fig:app_mockup}
\end{figure}

Figure \ref{fig:app_mockup} illustrates a possible application scenario that takes advantage of the ability to distinguish left and right hands. One could imagine that placing the non-dominant hand on the tabletop triggers the display of a menu where menu items are aligned with the fingers. Tapping a finger would either execute the corresponding action or display a submenu that again is aligned with the current finger positions. Hence the non-dominant hand is used for navigational tasks and mode-switching while the dominant hand is used for tasks that require particular precision. Furthermore that scenario would suit both novice and expert users. Novice users would benefit from the visual feedback provided by the menu display while expert users would only need to execute a tap sequence. Moreover as the non-dominant hand could be placed anywhere on the table, expert users could perform tasks without even looking at their hand during interaction.

